\documentclass[preprint]{imsart}

\usepackage{natbib}
\usepackage{hyperref}

\usepackage{graphicx} 
\usepackage[justification=centering]{subfig}
\usepackage{morefloats}
\usepackage{amsmath,amssymb} 
\usepackage{siunitx}
\usepackage{multirow}
\usepackage{booktabs}
\usepackage{color}
\usepackage{array}
\usepackage{tabularx}
\newcolumntype{Y}{>{\centering\arraybackslash}X}
\usepackage{dcolumn}
\newcolumntype{d}[1]{D{.}{.}{#1}}
\newcolumntype{.}{D{.}{.}{-1}}

\usepackage[table]{xcolor}
\definecolor{scoreAcolor}{rgb}{1,0.9,0.9}
\definecolor{scoreBcolor}{rgb}{0.9,1,0.9}
\definecolor{scoreCcolor}{rgb}{0.9,0.9,1}
\definecolor{scoreDcolor}{rgb}{1,1,0.9}
\newcommand{\scoreA}[1]{\cellcolor{scoreAcolor} {#1}}
\newcommand{\scoreB}[1]{\cellcolor{scoreBcolor} {#1}}
\newcommand{\scoreC}[1]{\cellcolor{scoreCcolor} {#1}}
\newcommand{\scoreD}[1]{\cellcolor{scoreDcolor} {#1}}

\newcommand{\set}[1]{{\mathcal N}}

\begin{document}

%
%
%
%

\begin{frontmatter}
\title{Computer vision tools for the non-invasive assessment of autism-related behavioral markers\thanksref{t2}}
\runtitle{Computer vision tools for assessing ASD behavioral markers}

  \thankstext{t2}{J.~Hashemi, T.~V.~Spina, and M.~Tepper equally contributed to this work. This work was partially done while M.~Tepper, T.~V.~Spina,  and G.~Sapiro were with the University of Minnesota.
  Work partially supported by NSF Grants 1039741 \& 1028076, CAPES (BEX 1018/11-6) and FAPESP (2011/01434-9) PhD scholarships from Brazil, and the U.S. Department of Defense.}

\begin{aug}
  \author{
    \fnms{Jordan} \snm{Hashemi}%
    \corref{}%
    \ead[label=e1]{hashe007@umn.edu}
  }

  \address{
    Department of Electrical and Computer Engineering, University of Minnesota, USA.\\
    \printead{e1}
  }

  \author{
    \fnms{Thiago} \snm{Vallin Spina}%
    \corref{}%
    \ead[label=e2]{tvspina@ic.unicamp.br}
  }

  \address{
    Institute of Computing, University of Campinas, Brazil\\
    \printead{e2}
  }

  \author{
    \fnms{Mariano} \snm{Tepper}%
    \corref{}%
    \ead[label=e3]{mariano.tepper@duke.edu}
  }

  \address{
    Department of Electrical and Computer Engineering, Duke University, USA.\\
    \printead{e3}
  }

  \author{
    \fnms{Amy} \snm{Esler}
  }

  \address{
    Department of Pediatrics, University of Minnesota, USA.
  }

  \author{
    \fnms{Vassilios} \snm{Morellas}
  }

  \author{
    \fnms{Nikolaos} \snm{Papanikolopoulos}
  }

  \address{
    Department of Computer Science and Engineering, University of Minnesota, USA.
  }

  \author{
    \fnms{Guillermo} \snm{Sapiro}
  }

  \address{
    Department of Electrical and Computer Engineering and Department of Computer Science, Duke University, USA.
  }

  \runauthor{J. Hashemi et al.}

\end{aug}

\begin{abstract}
The early detection of developmental disorders is key to child
outcome, allowing interventions to be initiated that promote
development and improve prognosis. Research on autism spectrum
disorder (ASD) suggests behavioral markers can be observed late in the
first year of life. Many of these studies involved extensive
frame-by-frame video observation and analysis of a child's natural behavior. Although non-intrusive, these methods are
extremely time-intensive and require a high level of observer
training; thus, they are impractical for clinical and large population research purposes. Diagnostic
measures for ASD are available for infants but are only accurate when
used by specialists experienced in early diagnosis. This work
is a first milestone in a long-term multidisciplinary project that
aims at helping clinicians and general practitioners accomplish this
early detection/measurement task automatically. We focus on providing
computer vision tools to measure and identify ASD behavioral markers
based on components of the Autism Observation Scale for
Infants (AOSI). In particular, we develop algorithms to measure three
critical AOSI activities that assess visual attention. We augment
these AOSI activities with an additional test that analyzes
asymmetrical patterns in unsupported gait. The first set of algorithms
involves assessing head motion by tracking facial features, while the
gait analysis relies on joint foreground segmentation and 2D body pose
estimation in video. We show results that provide insightful knowledge
to augment the clinician's behavioral observations obtained from real
in-clinic assessments.
\end{abstract}


\end{frontmatter}

\section{Introduction}

The analysis of children's natural behavior is of key importance for
the early detection of developmental disorders such as autism spectrum
disorder (ASD). For example, several studies have revealed behaviors
indicative of ASD in early home videos of children that were later
diagnosed with ASD~\citep[see][and references
therein]{zwaigenbaum05}. These studies involved video-recording these environments
and then analyzing the data a posteriori, using frame-by-frame
viewing by an observer who typically trains for several weeks to
achieve inter-rater reliability. Of course, hours and hours of labor
are required, making such analyses impractical for clinical
settings as well as for big data studies aiming at the discovery or improvement of behavioral markers
While clinical tools for early diagnosis of ASD are available, they
require administration and interpretation by specialists. Most families lack easy access to specialists in
ASD; for example, the wait list for an evaluation at the leading ASD
Clinic at the University of Minnesota is 6 months for children age 4 and
under. There is a need for automatic and quantitative analysis tools
that can be used by general practitioners in child development, and in
general environments, to identify children at-risk for ASD and other
developmental disorders.


As a first milestone in this long-term goal, this work focuses on
providing computer vision tools for aiding in-clinic early diagnosis
of ASD. Although much is unknown about the
underlying causes of ASD, it is characterized by abnormalities in social interactions
and communication and the presence of restricted, repetitive
behaviors~\citep{zwaigenbaum05}. Neuropathological studies indicate that ASD has its origins in abnormal brain development early in prenatal life~\citep{rodier2002}.
Moreover, \citet{zwaigenbaum05} argue that many children with ASD exhibit several specific behavioral markers as early as in the first year of life. These markers appear, among others, in activities involving visual attention, often expressed as difficulties in disengagement and shifting of attention \citep{landry2004imp}. Once they begin walking, many children also show atypical motor patterns, such as asymmetric gait or toe walking \citep{Esposito11}.

Despite this evidence, the average age of ASD diagnosis in the US is 5
years~\citep{Shattuck09}. Recently, much research and clinical trials
have focused on early diagnosis to allow for early intensive
intervention. Early intervention, initiated in preschool and sustained
for at least 2 years, can substantially improve child outcomes~\citep[e.g.,][]{Dawson08}.
Detecting ASD risk and starting interventions
before the full set of behavioral symptoms appears has an even greater
impact, preventing difficult behaviors and delayed developmental
trajectories from taking hold~\citep{Dawson08}. Early diagnosis is
achieved by following a comprehensive  battery of developmental and behavioral tests and parent interviews, with the goal of detecting behavioral symptoms consistent with ASD. However, few specialized clinics exist to offer these assessments to the very young. In the US, the average age of diagnosis is 5 years~\citep{Shattuck09}. Improving availability of early diagnosis may be achieved by developing screening tools that can be used in regular pediatric clinics and school environments, thereby reaching a large population very early. Towards this end, in this work, we develop semi-automatic computer vision video analysis techniques to aid in early detection.



These tools aid the practitioner in the diagnosis task by providing
accurate and objective measurements.  In addition, and particularly
for research, automatic analysis will permit to analyze effortlessly
vast amounts of naturally recorded videos, opening the door for data
mining towards the improvement of current assessment protocols and the
discovery of new behavioral features.  This project is being developed
by a multidisciplinary group bringing together professionals from
psychology, computer vision, and machine learning. As opposed to other
research projects~\citep{jones,chapman,Klin}, where artificial setups
are used, one of our main goals is to provide non-intrusive capturing
systems that do not necessarily induce behavioral modification in the children. In
other words, hardware must not constrain the testing environment: the
clinician is free to adjust testing conditions as needed, and children
are not asked to wear any type of sensors~\citep{Goodwin11,Nazneen10}
or perform any non-natural tasks.

The results in this paper are from actual clinical recordings, in
which the at-risk infant/toddler is tested by an experienced clinician
following the Autism Observation Scale for Infants
(AOSI)~\citep{bryson2007prospective} and a standard battery of
developmental and ASD assessment measures (e.g., the Autism Diagnostic
Observation Schedule -- Toddler Module, ADOS-T,~\citet{Luyster09}; and
the Mullen Scales of Early Learning, MSEL,~\citet{Mullen95}). The AOSI
is a well-validated behavioral observation tool~\citep{Downing11} for
gathering information on early ASD risk signs, involving a set of
semi-structured activities that provide an interactive context in
which the examiner engages the infant in play, while conducting a set
of systematic presses to elicit specific child behaviors. In our
clinical setup, we use two low-cost GoPro Hero HD color cameras (with
a resolution of 1080p at 30 fps), one placed on the clinician's table
(e.g., Figure~\ref{f.shared_interest}) and one in a corner of the room
(Figure~\ref{f.pose-estimation}); the displayed images are here
downsampled, blurred, and/or partially blocked to preserve anonymity
(processing was done on the original videos).\footnote{Approval for
  this study was obtained from the Institutional Review Board at the
  University of Minnesota.}

We present video analysis tools for assessing four fundamental
behavioral patterns: visual tracking, disengagement of attention,
sharing interest, and atypical motor behavior (full session). The
first three are part of the AOSI while the latter is holistically
assessed throughout the whole session.  The first three behaviors will
be addessed by tracking simple facial features and estimating the head
movements from them. The last behavior is treated using a joint body
segmentation/pose estimation algorithm.  The work with such specific
population of infants and toddlers is unique in the computer vision
community, making this a novel application for the psychology
community. While the data is obtained from actual clinical assessments, the tasks pulled from the assessment are easy to administer and/or involve recordings of the child's natural motor behavior, opening the door to broad behavioral studies, considering that the actual analysis is automatically done as here introduced.

In the following sections, we first describe our proposed automatic head pose
tracking and body pose estimation methods. Afterwards, we detail our
experimental validation that involved comparing our results with the
clinician's scores of the evaluation session as well as with non-expert manual scoring.
\section{Assessing Visual Attention}
\label{sec:attention}

Through the development of the AOSI, \citet{zwaigenbaum05} identified multiple behavioral markers for
early detection of ASD. We focus on three of these, namely sharing
interest, visual tracking, and disengagement of attention. The AOSI
states specific guidelines on how to evaluate these behavioral markers
from their corresponding activities.

The AOSI is divided into two main parts/categories: (1) a set of tabulated tasks which are designed for assessing specific behaviors; each task consists of  a certain number of presses and the child's responses receive a score; (2) a freeplay session, in which the clinician assesses the social behavior of the child while he is allowed to explore toys/objects.
In this work we focus on computer vision tools for two AOSI tasks which belong to the first category.

\noindent\textbf{Visual Tracking.}
It represents the ``ability to visually follow a moving object laterally across
the midline''~\citep{bryson2007prospective}. To evaluate it,
the following activity is performed: (1) a rattle or other noisy toy
is used to engage the infant's attention, (2) the rattle is
positioned to one side of the infant, and (3) the rattle is then moved
silently at eye level across the midline to the other side (note the relative simplicity of administering this and the additional tasks described next). 
The clinician evaluates how well the infant tracks the moving object. Infants with
ASD usually exhibit discontinuous and/or a noticeably delayed tracking~\citep{bryson2007prospective}.

\noindent\textbf{Disengagement of Attention.}
It is characterized as the ``ability to disengage and move
eyes/attention from one of two competing visual
stimuli''~\citep{bryson2007prospective}. The corresponding activity
consists of (1) shaking a noisy toy to one side of the infant until
his/her attention is engaged, and (2) then shaking a second noisy toy
on the opposite side, while continuing to shake the first object.
The clinician assesses the child's ability to shift attention away from
one object when another is presented. A delayed response is
an ASD risk sign~\citep{landry2004imp}.

Throughout the freeplay session, the clinician extracts many behavioral measurements. The studied behaviors mainly include social interactions and thus their assessments also take place throughout the entire session. The less structured nature of these holistic assessments makes the development of automated tools harder. We will show, however, that the computer vision tools presented in this work can also be of use in a more complex scenario, such as the freeplay session. We then explore in detail one activity belonging to the freeplay session, the ball playing activity, as a first example of the potential uses of our approach.

\noindent\textbf{Sharing Interest.}  It is described as the ``ability
to use eyes to reference and share interest in an object or event with
another person'' (\citet{bryson2007prospective}, also known as "Social
interest and shared affect"). Although this behavior is evaluated
throughout the AOSI, it can be specifically assessed from a ball
playing activity, in which a ball is rolled on the table towards the
infant after engaging his/her attention. After receiving the ball, the
clinician analyzes the child's ability to acknowledge the involvement
of another person in the gameplay by looking to either the clinician
or the caregiver. Infrequent or limited looking to faces is an early ASD
risk sign~\citep{zwaigenbaum05,bryson2007prospective}.


To analyze the child's reactions in the Visual Attention activities, we automatically estimate the changes of two head pose motions: yaw (left and right motion) and pitch (up and down motion). For the Visual Tracking and Disengagement of Attention tasks, which involve lateral motions, we focus on the yaw motion; conversely, in the Sharing Interest task, we mainly focus on the pitch motion.
We present computer vision algorithms for estimating these head motions. The algorithms track specific facial features: the left ear, left eye, and nose, see for example Figure~\ref{f.triangle}. From their positions we compute an estimate of the participant's yaw and pitch motions.
The only user input in our algorithm is during initialization. On the first frame, the user places a bounding box around the left ear, left eye, and nose. This could potentially be avoided by standard feature detection techniques.
We marked the playing objects by hand, although this also can be done automatically from prior knowledge of their visual and sound features (e.g., color or squeaking noise).
Additional technical details are available in Appendix~\ref{app:tracking}.


\section{Assessing Motor Patterns}
\label{sec:motor}

\newcommand{\ASa}[0]{\ensuremath{AS^*}}
\newcommand{\ASu}[0]{\ensuremath{AS_u}}
\newcommand{\ASf}[0]{\ensuremath{AS_f}}
\newcommand{\ADf}[0]{\ensuremath{AD_f}}

Motor development has often been hypothesized as an early bio-marker
of autism, and motor development disorders are considered some of the
first signs which could precede social or linguistic
abnormalities~\citep[and references therein]{Esposito11}. Hence, it is
important to find means of detecting and measuring these atypical
motor patterns at a very early stage. In the AOSI protocol, atypical
motor behavior is portrayed as the ``presence of developmentally
atypical gait, locomotion, motor mannerisms/postures or repetitive
motor behaviours''~\citep{bryson2007prospective}. There is no specific
activity for assessing motor patterns; the clinician performs a
holistic evaluation of the behaviors by visual inspection whenever
they occur throughout the full
session~\citep{Mullen95,bryson2007prospective}.

Children diagnosed with autism may present arm-and-hand flapping, toe
walking, asymmetric gait when walking unsupportedly, among other
atypical motor behaviors. In particular, \citet{Esposito11} have found
that diagnosed toddlers often presented asymmetric arm positions
(Figure~\ref{f.p19-s01-asymmetry}), according to the Eshkol-Wachman
Movement Notation (EWMN)~\citep{Teitelbaum04}, in home videos filmed
during the children's early life period. EWMN is essentially a 2D
stickman that is manually adjusted to the child's body on each video
frame and then analyzed. Our goal is to semi-automate this task by
estimating the 2D body pose of the toddlers in video segments in which
they are walking naturally.

Human body pose estimation is a complex and relatively well explored
research topic in computer vision~\citep{Kohli08,Ionescu11,Eichner12},
although it has been mostly restricted to adults, often in constrained
scenarios, and not yet exploited in the application we address. We
approach 2D human pose estimation task by using an extension of the
Object Cloud Model (OCM) segmentation framework that works with
articulated structures and video data.\footnote{The CSM extension was jointly developed with Alexandre X. Falc\~{a}o} Additional technical details are
available in Appendix~\ref{app:bodyPose}. 
Once the skeleton is estimated for each video
segment frame, we may extract angle measures to estimate arm
asymmetry. In this work, we treat the arm asymmetry estimation as an
application for the 2D body pose estimation, while hypothesizing that
action recognition methods based on body pose and/or point trajectory
estimation~\citep{Yao12,Sivalingam12} might be further applied to
automatically detect and measure other important stereotypical motor
behaviors (e.g., arms parallel to the ground pointing forward,
arm-and-hand flapping).

\subsection{Arm Asymmetry Measurement From 2D Body Pose}

\begin{figure}
\begin{center}
\begin{tabular}{cc}
\includegraphics[width=.2\textwidth]{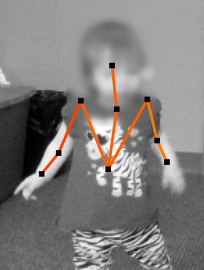} &
\includegraphics[width=.2\textwidth]{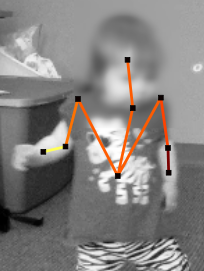}
\end{tabular} 
\caption{Example of symmetric and asymmetric arms. The sticks (skeleton) are automatically positioned with the technique here developed.
\label{f.asymmetric-arm}}
\end{center}
\end{figure}

Following~\citet{Esposito11}, a symmetrical position of the arms is a
pose where similarity in relative position of corresponding limbs (the
left and right arms) is shown with an accuracy of $45^o$. This happens
because EWMN defines a 3D coordinate system for each body joint that
discretizes possible 2D skeleton poses by equally dividing the 3D
space centered at the joints into $45^o$ intervals. Symmetry is
violated, for example, when the toddler walks with one arm fully
extended downwards alongside his/her body, while holding the other one
horizontally, pointing forward (Figure~\ref{f.asymmetric-arm}).

In our dataset, we have observed that using simple measures obtained
directly from the automatically computed 2D skeleton is often insightful enough to detect
most cases of arm asymmetry, thus avoiding the manual annotation
required by EWMN according to the aforementioned coordinate
system. For such asymmetry detection task, we define two
scores: \ASa{} and \ADf{}. \ASa{} is a normalized asymmetry score
(ranging between $[0,2.0]$) that takes into account both global and
relative angles from the skeleton arm segments (forearm and upper
arm). \ADf{} is based on the difference between the left and right
forearms global angles with respect to the horizontal axis, ranging
between $[0,180^o]$~\citep{Hashemi12}. These measures indicate arm
asymmetry when either $\ASa\geq 1.0$ or $\ADf\geq45^o$. Both \ASa{}
and \ADf{} have different advantages and shortcomings that will be
discussed in the experimental validation section. Nevertheless, \ASa{}
is the standard measure we adopt for most of our results. See
Appendix~\ref{ss.asymmetry-measures} for more details on how to
compute \ASa{} and \ADf{}.

\newcommand{\pnineteen}{\ensuremath{\#1}}
\newcommand{\pfive}{\ensuremath{\#2}}
\newcommand{\ptwo}[0]{\ensuremath{\#3}}
\newcommand{\pthree}[0]{\ensuremath{\#4}}
\newcommand{\pone}[0]{\ensuremath{\#5}}
\newcommand{\pseven}[0]{\ensuremath{\#6}}

\newcommand{\psix}{\ensuremath{\#7}}
\newcommand{\peight}{\ensuremath{\#8}}
\newcommand{\pnine}{\ensuremath{\#9}}
\newcommand{\ptwelve}{\ensuremath{\#10}}
\newcommand{\pfourteen}{\ensuremath{\#11}}
\newcommand{\ptwenty}{\ensuremath{\#12}}
\newcommand{\pfour}{\ensuremath{\#13}}
\newcommand{\pfiveone}{\ensuremath{\#14}}
\newcommand{\psevenone}{\ensuremath{\#15}}

\section{Experimental Validation}

This study involves 15 participants,\footnote{Participants \pfive{} and
  \pfiveone, \pseven{} and \psevenone{} are the same. However, the
  videos we use for \pfiveone{} and \psevenone{} are from their first
  evaluation sessions, while the ones we use for \pfive{} and
  \pseven{} are from their second sessions (when they reached walking
  age).} including both males and females ranging in
age from 5 to 18 months. All participants were classified as a baby
sibling of someone with ASD, a premature infant, or as a participant
showing developmental delays. Table~\ref{part_info} presents a summary of this information.
Note that, the participants are not clinically diagnosed until they are $36$ months of age and only
participant \ptwo{} has presented conclusive signs of ASD.

\begin{table}
\caption{Information on Participants involved in this study. Each participant was chosen for a different reason: being a baby sibling of someone with ASD, a premature infant, or showing developmental delays.}

\centering
\begin{tabular}{cccc}
\toprule

Part \# & Age (months) & Gender & Risk Degree \\ 
\midrule
\pnineteen & 14 & F & Showing delays\\
\pfive & 11 & M & Premature infant\\
\ptwo & 16 & M & ASD diagnosed\\
\pthree & 15 & M & Showing delays\\
\pone & 16 & M & Baby sibling\\
\pseven & 12 & F & Premature infant\\
\psix & 10 & F & Premature infant\\
\peight & 9 & M & Premature infant\\
\pnine & 7 & M & Premature infant\\
\ptwelve & 6 & M & Baby sibling\\
\pfourteen & 9 & M & Premature infant\\
\ptwenty & 18 & M & Showing delays\\
\pfour & 5 & F & Baby sibling\\
\pfiveone & 8 & M & Premature infant\\ 
\psevenone & 9 & F & Premature infant\\
\bottomrule
\end{tabular}
\label{part_info}

\end{table}

\subsection{Specific AOSI tasks}

During the AOSI assessment, the clinician performs three trials for the Disengagement of Attention task and two trials for the Visual Tracking task, per participant. Every trial receives an AOSI-tabulated score, according to the following guidelines:
\begin{itemize}
\item Disengagement of Attention. A trial is considered ``passed'' if the child looks to the second object in less than 1s, considered ``delayed'' if the child looks after a 1-2s delay, and considered ``stuck'' if the child looks after more than 2s.
\item Visual Tracking. During this task, the AOSI focuses on how smooth the participant is able to track the object. Depending on how continuously and smoothly the partipant is able to track the object, the trial is considered ``passed,'' ``delayed or interrupted,'' or ``partial or no tracking.''
\end{itemize}

The clinician makes a ``live'' judgment about these time frames or may
look at videos of this task if available. Finally, an overall score
for each task is computed by merging the individual ones. We followed
the protocol of comparing the assessments done by: (1) an expert
psychologist in autistic children, (2) a child/adolescent
psychiatrist, (3) two psychology students with no particular autism
training, and (4) the results of our computational tools. This setup
allows to contrast the automatic method's findings with human
assessments across the full range of expertise.

\begin{table}
\caption{Results of Disengagement of Attention task. A trial is
  considered either as ``passed'' (Pass), ``delayed'' (Del), or ``stuck'' (Stck) depending
  on whether the child disengages from the first object in less than
  $1s$, between $1-2s$, or more than $2s$, respectively. The proposed method emulates accurately the clinician's assessment (colors are added to facilitate the comparison). We also present the automatically computed delay that the child takes to disengage. Note that we consider a $+ \frac{1}{3}$ of a second margin for each delay to accommodate human error of making a live judgment.}

\begin{tabularx}{\textwidth}{cYY.YY.YY.}
\toprule
\multirow{3}{*}{Part.} & \multicolumn{3}{c}{First Trial Score} & \multicolumn{3}{c}{Second Trial Score} & \multicolumn{3}{c}{Third Trial Score} \\

\cmidrule(lr){2-4} \cmidrule(lr){5-7} \cmidrule(lr){8-10}
& \multirow{2}{*}{Clin.} & \multicolumn{2}{c}{Automatic} & \multirow{2}{*}{Clin.} & \multicolumn{2}{c}{Automatic} & \multirow{2}{*}{Clin.} & \multicolumn{2}{c}{Automatic} \\ 
\cmidrule(lr){3-4} \cmidrule(lr){6-7} \cmidrule(lr){9-10}
& & Score & \multicolumn{1}{c}{\text{Delay (s)}} & & Score & \multicolumn{1}{c}{\text{Delay (s)}} & & Score & \multicolumn{1}{c}{\text{Delay (s)}} \\
\midrule
\pnineteen & \scoreA{Pass} & \scoreA{Pass} & 0.9 & \scoreA{Pass} & \scoreA{Pass} & 0.7 & \scoreA{Pass} & \scoreA{Pass} & 0.37\\
\ptwo & \scoreA{Pass} & \scoreA{Pass} & 0.5 & - & - & \text{-} & - & - & \text{-}\\
\pthree & \scoreA{Pass} & \scoreA{Pass} & 0.23 & \scoreA{Pass} & \scoreA{Pass} & 1.1 & \scoreA{Pass} & \scoreA{Pass} & 0.3\\
\psix & \scoreA{Pass} & \scoreA{Pass} & 0.83 & \scoreA{Pass} & \scoreA{Pass} & 0.97 & \scoreA{Pass} & \scoreA{Pass} & 1.13\\
\peight & - & - & \text{-}  & \scoreA{Pass} & \scoreA{Pass} & 1.33 & - & - & \text{-} \\
\pnine & - & -  & \text{-}  & - & - & \text{-} & \scoreB{Del} & \scoreB{Del} & 1.37\\
\ptwelve & \scoreA{Pass} & \scoreA{Pass} & 0.87 & \scoreA{Pass} & \scoreA{Pass} & 1.3 & \scoreA{Pass} & \scoreA{Pass} & 1.33\\
\pfourteen & \scoreA{Pass} & \scoreA{Pass} & 0.83 & \scoreA{Pass} & \scoreA{Pass} & 0.63 & \scoreA{Pass} & \scoreA{Pass} & 0.87\\
\ptwenty & \scoreA{Pass} & \scoreA{Pass} & 0.93 & \scoreA{Pass} & \scoreA{Pass} & 0.9 & \scoreA{Pass} & \scoreA{Pass} & 0.87\\
\pfour & N/A & \scoreB{Del} & 1.87 & - & - & \text{-} & - & - & \text{-}\\
\pfiveone & \scoreB{Del} & \scoreA{Pass} & 1.07 & \scoreB{Del} & \scoreB{Del} & 1.77 & \scoreA{Pass} & \scoreA{Pass} & 0.5\\
\psevenone & \scoreA{Pass} & \scoreA{Pass} & 1.03 & \scoreA{Pass} & \scoreB{Del} & 1.43 & \scoreA{Pass} & \scoreA{Pass} & 0.7\\
\bottomrule
\end{tabularx}
\label{vis_disengage}

\end{table}

Table~\ref{vis_disengage} summarizes the results of our method and the clinical assessment for the Disengagement of Attention task. After marking when the second object is presented, our method is able to automatically determine the delay from when the participant disengages from the first object to the second. We present this delay in terms of how many seconds/frames it takes for the participant to disengage (note that we are recording the video at 30 frames per second). We incorporate a $ + \frac{1}{3}$ of a second margin for each delay to accommodate human error of making a live judgment. Out of the 24 trials that the clinician assigned a ``pass'' score to, our method agreed on 23 of them and scored a ``delayed'' for the other trial. And out of the 3 trials the clinican scored ``delayed'' our method agreed on 2 trials, scoring one as a ``pass.''  Although our method obtained one false positive by scoring one trial ``delayed'' which the clinician scored as ``pass'' and missed one ``delayed'' trial, we believe one of the greatest impacts of our method is that it gives the clinician quantifiable data for this task and may allow to readjust the rigid scoring intervals provided in the AOSI. With a study on a larger population, new time intervals (and their variability) for scoring may be discovered, and these false positives could be analyzed not as a strict ``pass'' or ``delayed'' but as something in between.

\begin{table}
\caption{Results of Visual Tracking task. A trial can be considered
  ``passed'' (Pass), ``delayed'' (Del), ``interrupted'' (Int), ``partial'' (Prt), or ``no
  tracking'' depending on how smoothly the child visually tracks the
  object. The proposed method emulates accurately the clinician's assessment (colors are added to facilitate the comparison).}

\centering
\begin{tabular}{c*{2}{c}@{\hspace{.3in}}*{2}{c}}
\toprule
\multirow{2}{*}{Part.} & \multicolumn{2}{c}{First Trial Score} & \multicolumn{2}{c}{Second Trial Score} \\

\cmidrule(lr{22pt}){2-3} \cmidrule(l{0pt}r){4-5} 
& Clinician & Automatic & Clinician & Automatic \\ 
\midrule
\pnineteen & \scoreA{Pass} & \scoreA{Pass} & \scoreA{Pass} & \scoreC{Int} \\
\ptwo & \scoreB{Del} & \scoreA{Pass} & \scoreA{Pass} & \scoreA{Pass} \\
\pthree & \scoreA{Pass} & \scoreA{Pass} & \scoreA{Pass} & \scoreA{Pass} \\
\psix & - & - & \scoreC{Int} & \scoreC{Int} \\
\peight & \scoreA{Pass} & \scoreA{Pass} & \scoreA{Pass} & \scoreA{Pass} \\
\pnine & \scoreA{Pass} & \scoreA{Pass} & \scoreA{Pass} & \scoreA{Pass} \\
\ptwelve & \scoreA{Pass} & \scoreA{Pass} & \scoreC{Int} & \scoreC{Int}  \\
\pfourteen & \scoreA{Pass} & \scoreA{Pass} & \scoreC{Int} & \scoreC{Int} \\
\ptwenty & \scoreA{Pass} & \scoreA{Pass} & \scoreA{Pass} & \scoreA{Pass} \\
\pfour & \scoreC{Int} & \scoreC{Int} & \scoreC{Int} & \scoreC{Int}  \\
\pfiveone & \scoreD{Prt} & \scoreC{Int} & \scoreD{Prt} & \scoreD{Prt}\\
\psevenone & \scoreA{Pass} & \scoreA{Pass} & - &  -\\

\bottomrule
\end{tabular}
\label{vis_tracking}
\end{table}

Table~\ref{vis_tracking} summarizes the results of our method and the
clinical assessment for the Visual Tracking task. The simple output of
our method allows to easily assess each trial by visual inspection and
score the trials as either ``pass,''  ``interrupted,''  ``partial,''  or
``no tracking.''  Examples of our method's measurements for a
``pass,''  ``interrupted,''  and ``partial'' tracking scores are
explained later in this section, see Figure~\ref{f.tracking} for a few
examples. Our results strongly correlate with that of the
clinician. Out of the 14 trials that the clinican assessed as
``pass,''  our method agreed with 13 of them and scored an
``interrupted'' for 1 of the trials. For all the 4 trials the
clinician assessed as ``interrupted,''  our automatic method was in
agreement. The clinician scored two trials as ``partial,''  our method
scored one of them as ``partial'' and the other as
``interrupted.''  Lastly, the clinician scored one trial as
``delayed;'' however, based on our non-intrusive camera placement (as selected by the practitioner), we
are not able to continuously extract the object's location accurately
enough to assign ``delayed'' scores. These results not only show a
strong correlation between the assessment of the clinican and our
method for the Visual Tracking task, but also provide the clinician
and future researchers accurate quantitative data.

The child/adolescent psychiatrist and two psychology students assigned their scores by following the AOSI guidelines, without prior training, while watching the videos used by the automatic method.
Their results (tables~\ref{t.students_disengagement} and~\ref{t.students_tracking}) not only illustrate the human training that needs to be done for these visual attention tasks but also the novelty of our method and its quantitative results. Out of the 27 Visual Disengagement trials, the two psychology students agreed with the clinician on 13 and 16 of the trials respectively, while the child/adolescent psychiatrist agreed on 22 trials. Similarly for the 22 Visual Tracking trials, the two psychology students agreed with the clinician on 13 and 14 of the trials respectively, while the child/adolescent psychiatrist agreed on 16 trials. Table~\ref{t.humanVSautomatic} provides a summary of these results. The benefits of the results obtained with our automatic method for head pose estimation are threefold. First, it provides accurate quantitative measurements for the AOSI tasks, improving the shareability of clinical records (while not compromising anonymity). Second, it can also prove beneficial in the discovery of new behavioral patterns by easily collecting large amounts of data and using data mining on them. Third, it increases the granularity of the analysis by providing results at a finer scale. In the following, we provide in-depth analysis of some trials, which are relevant to show the validity of this argumentation.

\begin{table}
\caption{Human Results for Disengagement of Attention task. A trial is
  considered either as ``passed'' (Pass), ``delayed'' (Del), or ``stuck'' (Stck) depending
  on whether the child disengages from the first object in less than
  $1s$, between $1-2s$, or more than $2s$, respectively.
  Comparison of the clinician's scores (Clin.), the child/adolescent psychiatrist's scores (Psy.), and the two psychology students' scores (St.~1 and 2). Colors are added to facilitate the comparison.}

\begin{tabularx}{\textwidth}{c*{4}{Y}@{\hspace{.3in}}*{4}{Y}@{\hspace{.3in}}*{4}{Y}}
\toprule
\multirow{2}{*}{Part.} & \multicolumn{4}{c}{First Trial Score} & \multicolumn{4}{c}{Second Trial Score} & \multicolumn{4}{c}{Third Trial Score} \\

\cmidrule(lr{15pt}){2-5} \cmidrule(l{0pt}r{15pt}){6-9} \cmidrule(l{0pt}){10-13}
& Clin. & Psy. &  St.~1 & St.~2 & Clin. & Psy. &  St.~1 & St.~2 & Clin. & Psy. &  St.~1 & St.~2 \\ 
\midrule
\pnineteen & \scoreA{Pass} & \scoreA{Pass} & \scoreB{Del} & \scoreB{Del} & \scoreA{Pass} & \scoreA{Pass} & \scoreA{Pass} & \scoreA{Pass} & \scoreA{Pass} & \scoreA{Pass} & \scoreA{Pass} & \scoreA{Pass}\\
\ptwo & \scoreA{Pass} & \scoreA{Pass} & \scoreB{Del} & \scoreB{Del} & - & - & - & - & - & - & - & - \\
\pthree & \scoreA{Pass} & \scoreA{Pass} & \scoreA{Pass} & \scoreA{Pass} & \scoreA{Pass} & \scoreA{Pass} & \scoreA{Pass} & \scoreA{Pass} & \scoreA{Pass} & \scoreA{Pass} & \scoreA{Pass} & \scoreA{Pass}\\
\psix & \scoreA{Pass} & \scoreA{Pass} & \scoreB{Del} & \scoreB{Del} & \scoreA{Pass} & \scoreB{Del} & \scoreB{Del} & \scoreB{Del} & \scoreA{Pass} & \scoreA{Pass} & \scoreB{Del} & \scoreB{Del}\\
\peight & - & - &-  & - & \scoreA{Pass} & \scoreA{Pass} & \scoreB{Del} & \scoreB{Del} & -  & - & - & - \\
\pnine & - & -  & - & - &-  & - & - & - & \scoreB{Del} & \scoreA{Pass} & \scoreB{Del} & \scoreC{Stck}\\
\ptwelve & \scoreA{Pass} & \scoreB{Del} & \scoreB{Del} & \scoreB{Del} & \scoreA{Pass} & \scoreA{Pass} & \scoreA{Pass} & \scoreA{Pass} & \scoreA{Pass} & \scoreB{Del} & \scoreB{Del} & \scoreB{Del}\\
\pfourteen & \scoreA{Pass} & \scoreA{Pass} & \scoreB{Del} & \scoreA{Pass} & \scoreA{Pass} & \scoreA{Pass} & \scoreA{Pass} & \scoreA{Pass} & \scoreA{Pass} & \scoreA{Pass} & \scoreA{Pass} & \scoreA{Pass}\\
\ptwenty & \scoreA{Pass} & \scoreA{Pass} & \scoreA{Pass} & \scoreA{Pass} & \scoreA{Pass} & \scoreA{Pass} & \scoreB{Del} & \scoreB{Del} & \scoreA{Pass} & \scoreA{Pass} & \scoreA{Pass} & \scoreA{Pass}\\
\pfiveone & \scoreB{Del} & \scoreB{Del} & \scoreB{Del} & \scoreB{Del} & \scoreB{Del} & \scoreB{Del} & \scoreC{Stck} & \scoreB{Del} & \scoreA{Pass} & \scoreA{Pass} & \scoreB{Del} & \scoreA{Pass}\\
\psevenone & \scoreA{Pass} & \scoreA{Pass} & \scoreB{Del} & \scoreA{Pass} & \scoreA{Pass} & \scoreA{Pass} & \scoreA{Pass} & \scoreA{Pass} & \scoreA{Pass} & \scoreB{Del} & \scoreB{Del} & \scoreB{Del}\\
\bottomrule
\end{tabularx}
\label{t.students_disengagement}

\end{table}

\begin{table}
\caption{Human Results for Visual Tracking task. A trial can be considered
  ``passed'' (Pass), ``delayed'' (Del), ``interrupted'' (Int), ``partial'' (Prt), or ``no
  tracking'' depending on how smoothly the child visually tracks the
  object. The proposed method emulates accurately the clinician's assessment. Comparison of the clinician's scores (Clin.), the child/adolescent psychiatrist's scores (Psy.), and the two psychology students' scores (St.~1 and 2). Colors are added to facilitate the comparison.}

\begin{tabularx}{\textwidth}{c*{4}{Y}@{\hspace{.3in}}*{4}{Y}}
\toprule
\multirow{2}{*}{Part.} & \multicolumn{4}{c}{First Trial Score} & \multicolumn{4}{c}{Second Trial Score} \\

\cmidrule(lr{15pt}){2-5} \cmidrule(l{0pt}r){6-9} 
& Clin. & Psy. & St.~1 & St.~2 & Clin. & Psy. & St.~2 & St.~3 \\ 
\midrule
\pnineteen & \scoreA{Pass} & \scoreA{Pass} & \scoreA{Pass} & \scoreA{Pass} & \scoreA{Pass} & \scoreC{Int} & \scoreC{Int} & \scoreC{Int}\\
\ptwo & \scoreB{Del} & \scoreA{Pass} & \scoreA{Pass} & \scoreA{Pass} & \scoreA{Pass} & \scoreA{Pass} & \scoreA{Pass} & \scoreA{Pass} \\
\pthree & \scoreA{Pass} & \scoreA{Pass} & \scoreA{Pass} & \scoreA{Pass} & \scoreA{Pass} & \scoreA{Pass} & \scoreA{Pass} & \scoreA{Pass} \\
\psix & - & - & - & - & \scoreC{Int} & \scoreC{Int} & \scoreD{Prt} & \scoreD{Prt} \\ 
\peight & \scoreA{Pass} & \scoreA{Pass} & \scoreB{Del} & \scoreA{Pass} & \scoreA{Pass} & \scoreA{Pass} & \scoreA{Pass} & \scoreA{Pass} \\
\pnine & \scoreA{Pass} & \scoreA{Pass} & \scoreB{Del} & \scoreA{Pass} & \scoreA{Pass} & \scoreA{Pass} & \scoreA{Pass} & \scoreA{Pass} \\
\ptwelve & \scoreA{Pass} & \scoreA{Pass} & \scoreA{Pass} & \scoreA{Pass} & \scoreC{Int} & \scoreC{Int} & \scoreD{Prt} & \scoreD{Prt}  \\
\pfourteen & \scoreA{Pass} & \scoreA{Pass} & \scoreA{Pass} & \scoreA{Pass} & \scoreC{Int} & \scoreC{Int} & \scoreD{Prt} & \scoreD{Prt} \\
\ptwenty & \scoreA{Pass} & \scoreA{Pass} & \scoreA{Pass} & \scoreA{Pass} & \scoreA{Pass} & \scoreA{Pass} & \scoreA{Pass} & \scoreA{Pass}\\
\pfour & \scoreC{Int} & \scoreA{Pass} & \scoreC{Int} & \scoreA{Pass} & \scoreC{Int} & \scoreA{Pass} & \scoreA{Pass} & \scoreA{Pass}  \\
\pfiveone & \scoreD{Prt} & \scoreA{Pass} & \scoreC{Int} & \scoreA{Pass} & \scoreD{Prt} & \scoreB{Del} & \scoreD{Prt} & \scoreD{Prt}\\
\psevenone & \scoreA{Pass} & \scoreA{Pass} & \scoreA{Pass} & \scoreA{Pass} &- & - & - & -\\

\bottomrule
\end{tabularx}
\label{t.students_tracking}
\end{table}

\begin{table}

\caption{Number of agreements with the Autism expert for each partitipant in the two visual attention tasks. See tables~\ref{vis_disengage}, \ref{vis_tracking}, \ref{t.students_disengagement}, \ref{t.students_tracking} for individualized results.}
\label{t.humanVSautomatic}

\centering
\begin{tabular}{lccccc}
\toprule
Task & Trials & Automatic & Psychiatrist & Student 1 & Student 2\\
\midrule
Disengagment & 27 & \textbf{25} & 22 & 13 & 16 \\
Tracking & 22 & \textbf{19} & 16 & 13 & 14 \\
\cmidrule(rl){1-6}
Total & 49 & \textbf{44} & 38 & 26 & 30 \\
\bottomrule
\end{tabular}

\end{table}


Figure~\ref{f.disengagement} displays three important types of results for the Disengagement of Attention task.
In the first example, the participant is able to disengage from the first object and look at the second within 0.7s (21 frames) of the second object being presented. This would be scored as ``passed'' on the AOSI test. The participant in the second example disengages to the second object within 1.3s (40 frames), which would be scored as ``delayed'' on the AOSI test. The third example provides an interesting pattern in the participant's head movement. Not only does it take the third participant over 1s to look at the second object (which is ``delayed'' on the AOSI), but the participant displays piece-wise constant lateral head movements compared to the other two examples (which presented a much smoother motion), a pattern virtually impossible to detect with the naked eye. Again, such automatic and quantitative measurements are critical for aiding current and future diagnosis.

\begin{figure*}[b]
\centering
\includegraphics[width=.85\textwidth]{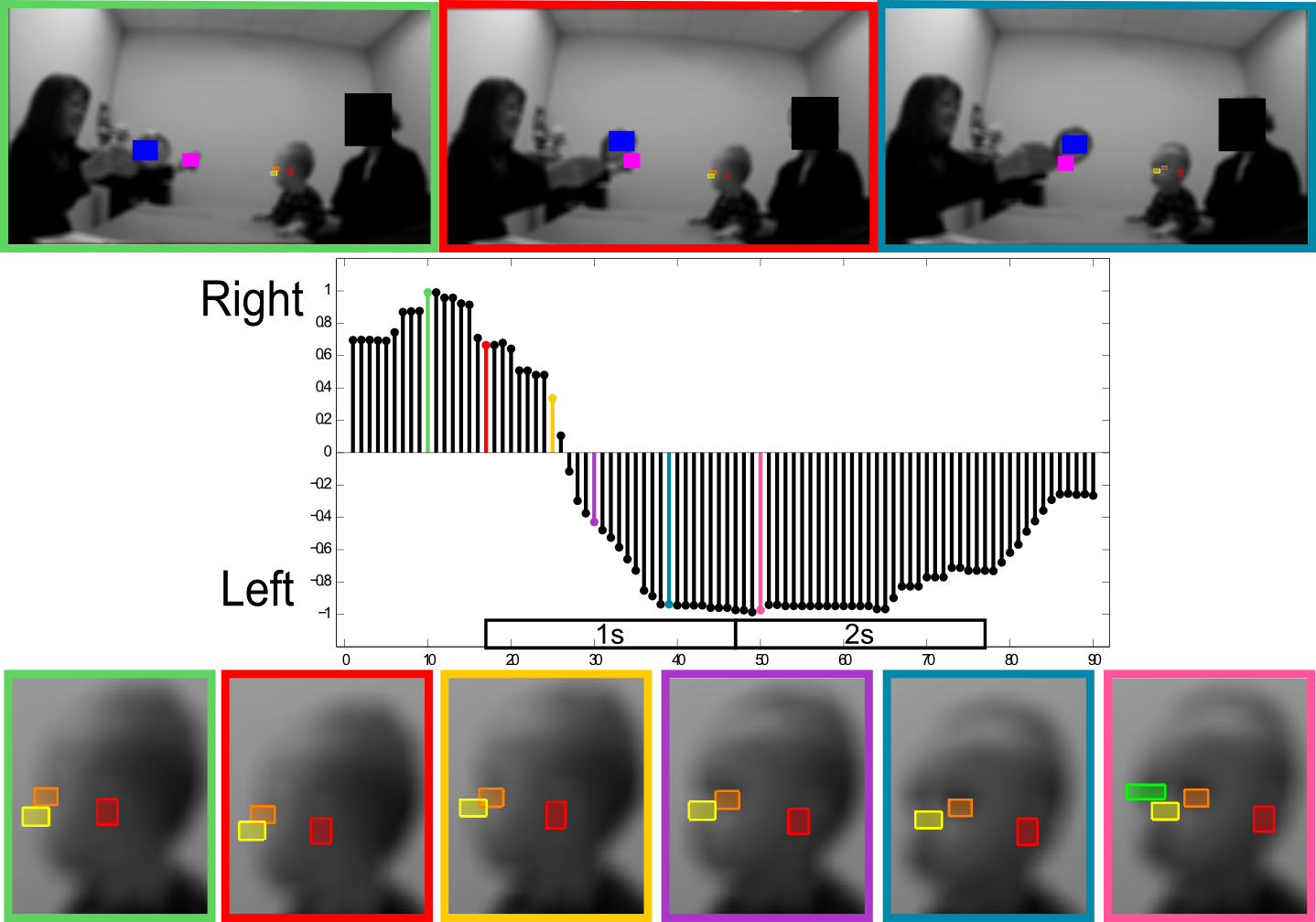}

\caption{Disengagement of Attention task. \textbf{Top:} clinician holding one object, when the second
  object appears, and when the child recognizes the second object.
  \textbf{Middle:} changes in the yaw motion ($\widehat{\mathrm{yaw}}$ values in the $y$-axis) for every
  frame ($x$-axis). The dotted line represents when the second object is presented, followed by boxes representing 1 and 2 seconds after the object is presented. \textbf{Bottom:} 6 examples of the infant's face
  during the task.
  All facial features are automatically detected and tracked (as indicated by the colored boxes around the nose, eyes and ear).
  Colors identify corresponding images and spikes in the graph.
\label{f.disengagement}}

\end{figure*}

\begin{figure*}[b]
\ContinuedFloat
\centering

\includegraphics[width=.85\textwidth]{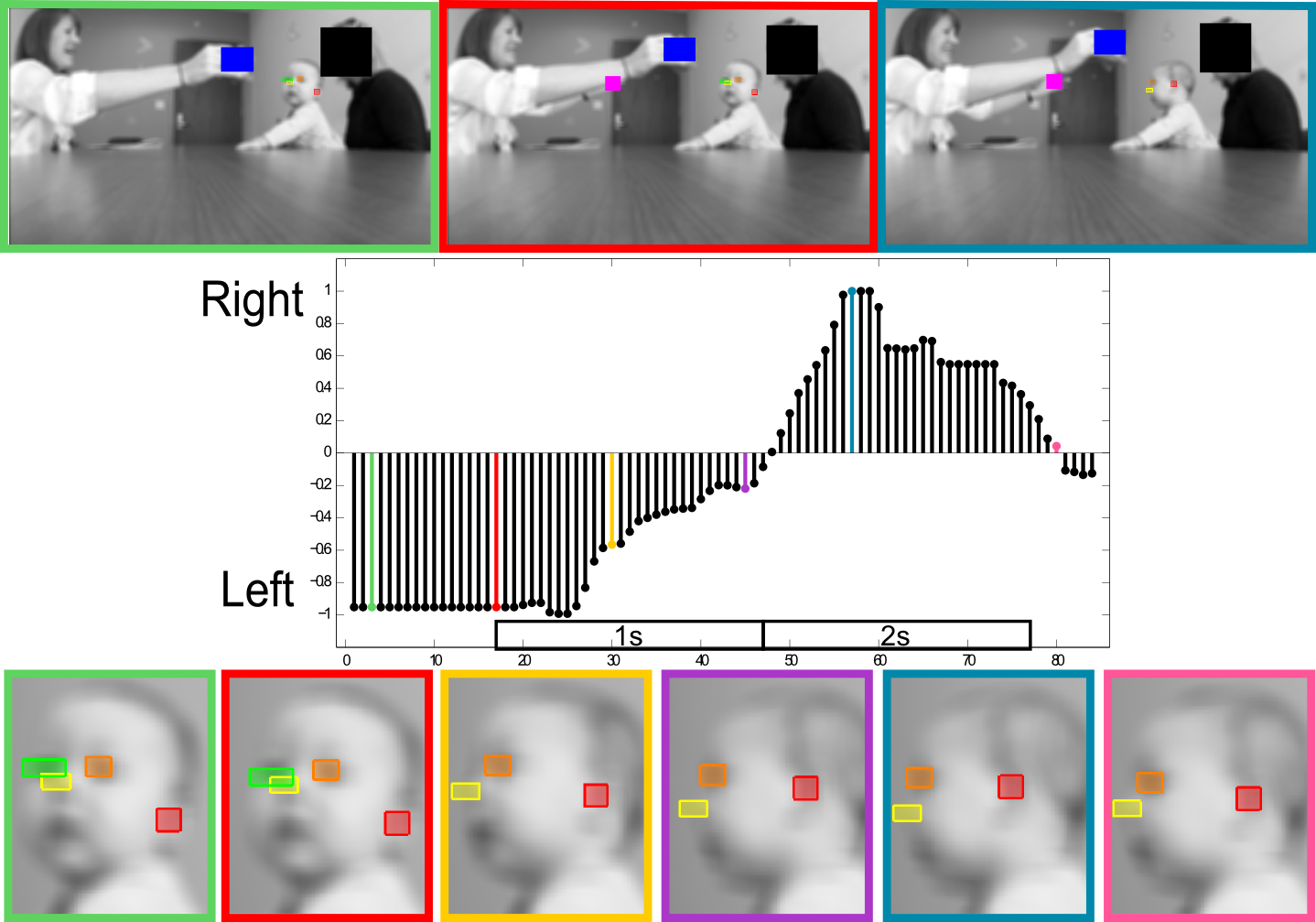}

\vskip0.5cm

\includegraphics[width=.85\textwidth]{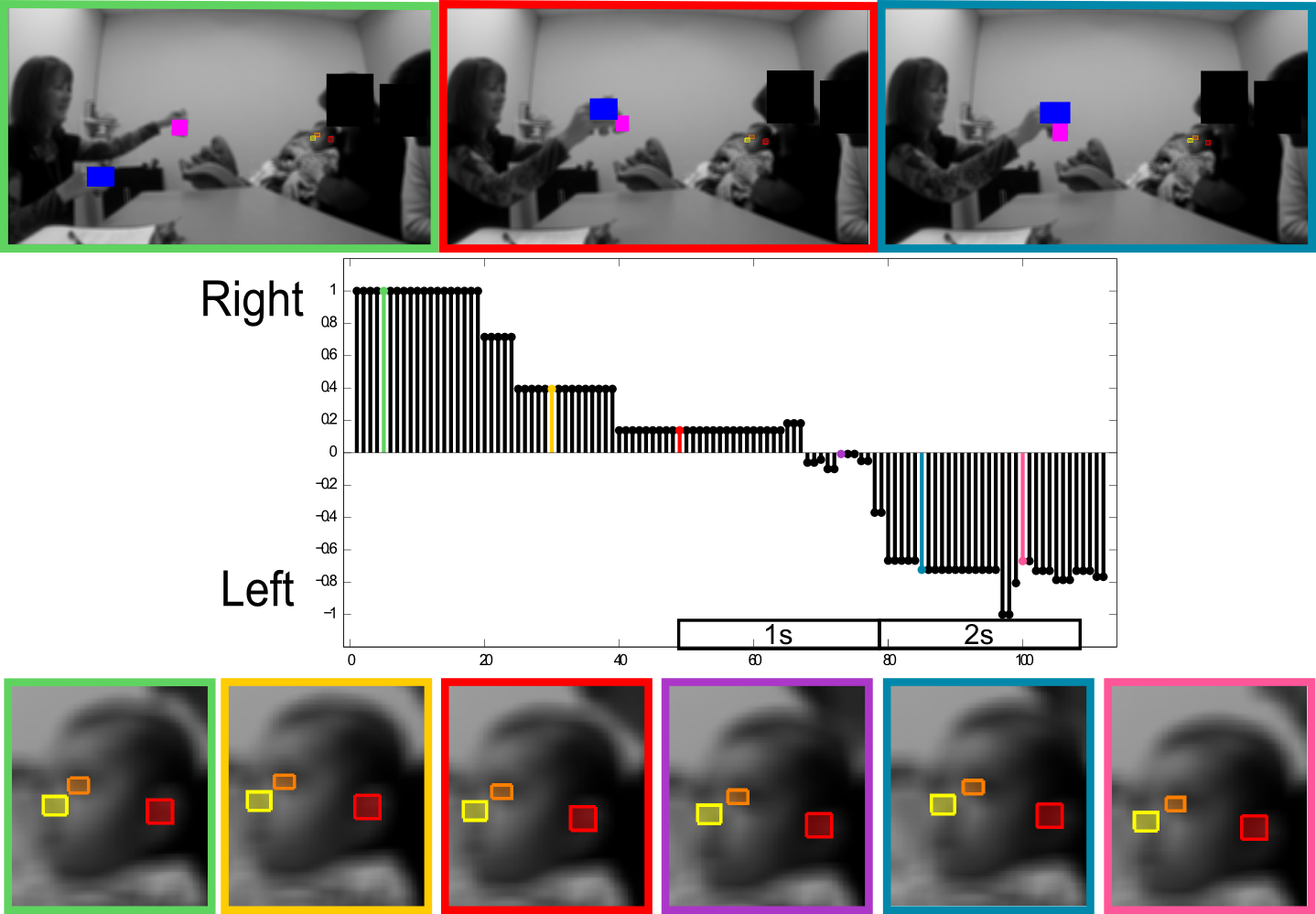}

\caption{(Continued) Disengagement of Attention task. \label{f.disengagement}
}
\end{figure*}

Figure~\ref{f.tracking} shows
three important types of our results for the Visual Tracking
task.  Not only does our method provide quantitative measurements for the Visual Tracking task, it also records the delay from when the object is at the second extreme side to when the participant's head is facing this side. 
The first example demonstrates a participant that received a ``passed'' on the AOSI's Visual Tracking task, since the participant was able to smoothly track the object with minimal delay as the object approached the participant's right. In the second example, the participant exhibited ``interrupted'' tracking motion. The participant's tracking of the object was interrupted as the object moved across the clinician's face. Instead of tracking the object as it moved across the clinician's face, the participant stopped tracking the object and looked at the clinician for 0.46s (14 frames) before continuing to track the object as it moved to the participant's left. Another aspect of our method is that it provides accurate and quantitative measurements for the participant's head tracking, thus one is able to automatically determine the delays between when the participant looks at the object or how long the participant stops his/her tracking. In the third example, the participant displays a ``partial'' tracking score on the AOSI. As the object crosses the clinician's face, the participant completely stops tracking the object and instead looks straight at the clinician.

\begin{figure*}[b]
\centering
\includegraphics[width=.85\textwidth]{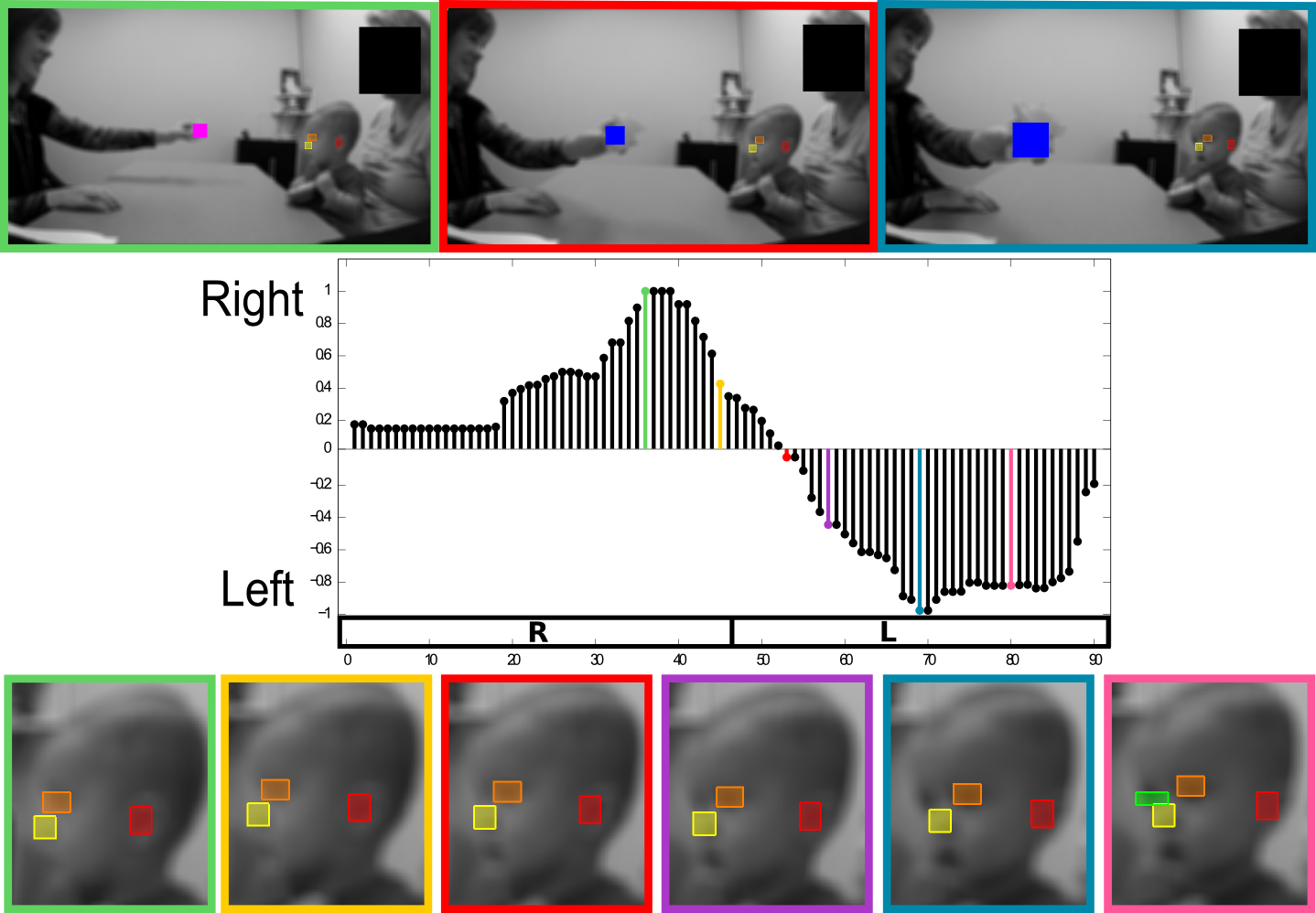}

\caption{Visual Tracking task. \textbf{Top:} the clinician holding the object,
  when the object is at one extreme side (right or left), and when the
  object is at the other extreme side.
  \textbf{Middle:} changes in the yaw motion ($\widehat{\mathrm{yaw}}$ values in the $y$-axis) for every
  frame ($x$-axis). The boxes labeled `R' and `L' represent when the object is to the right and left of the participant respectively. The gray shaded areas represent when the object is not moving and at an extreme side (either right or left). \textbf{Bottom:} 6 examples of the infant's face
  during the task. Colors identify corresponding images and spikes in the graph.
\label{f.tracking}}

\end{figure*}

\begin{figure*}
\ContinuedFloat
\centering

\includegraphics[width=.85\textwidth]{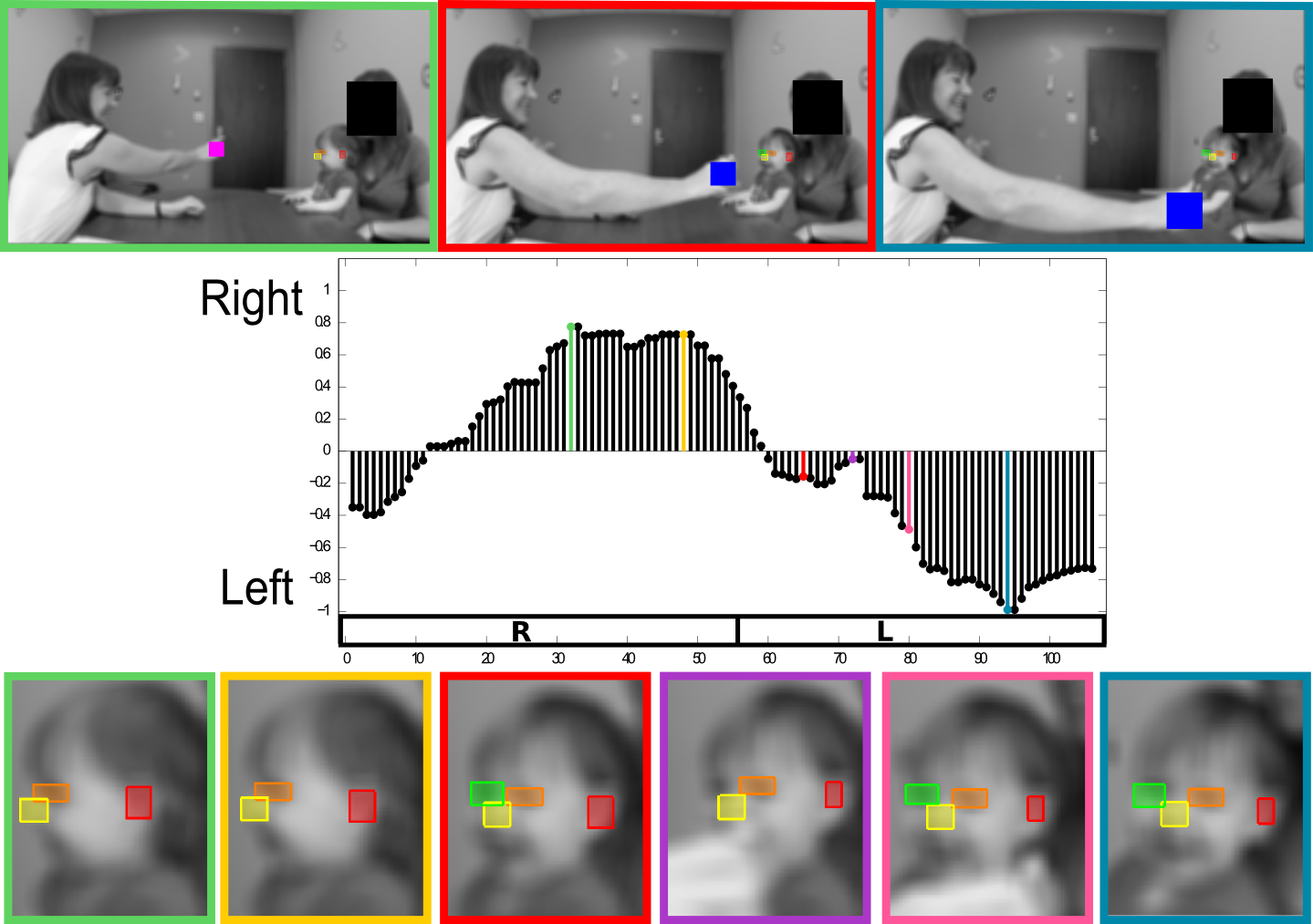}

\vskip0.5cm

\includegraphics[width=.85\textwidth]{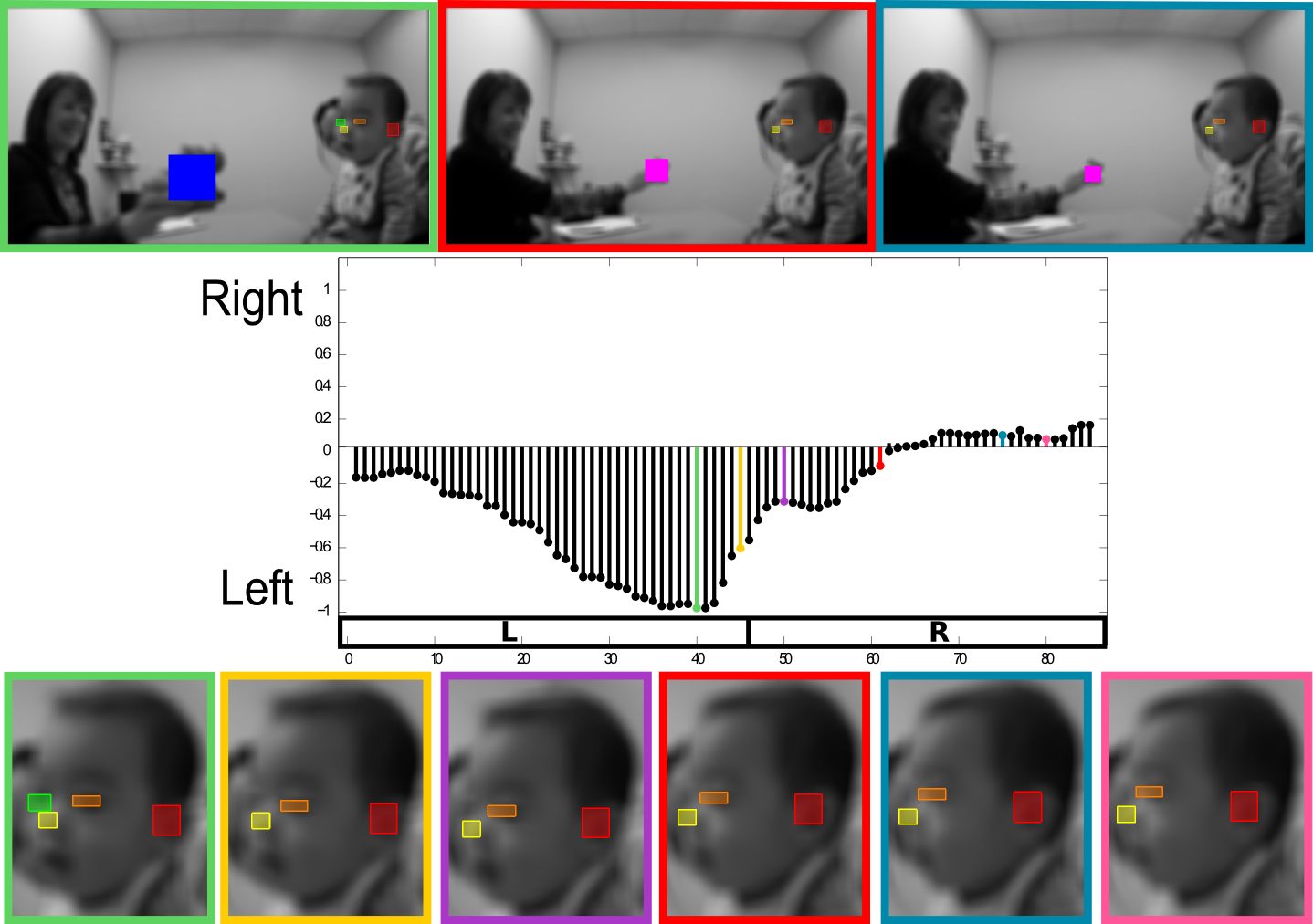}

\caption{(Continued) Visual Tracking task.\label{f.tracking}}
\end{figure*}

\subsection{The Ball Playing Activity During the Freeplay Session}

\begin{figure*}[b]
\centering
\includegraphics[width=.8\textwidth]{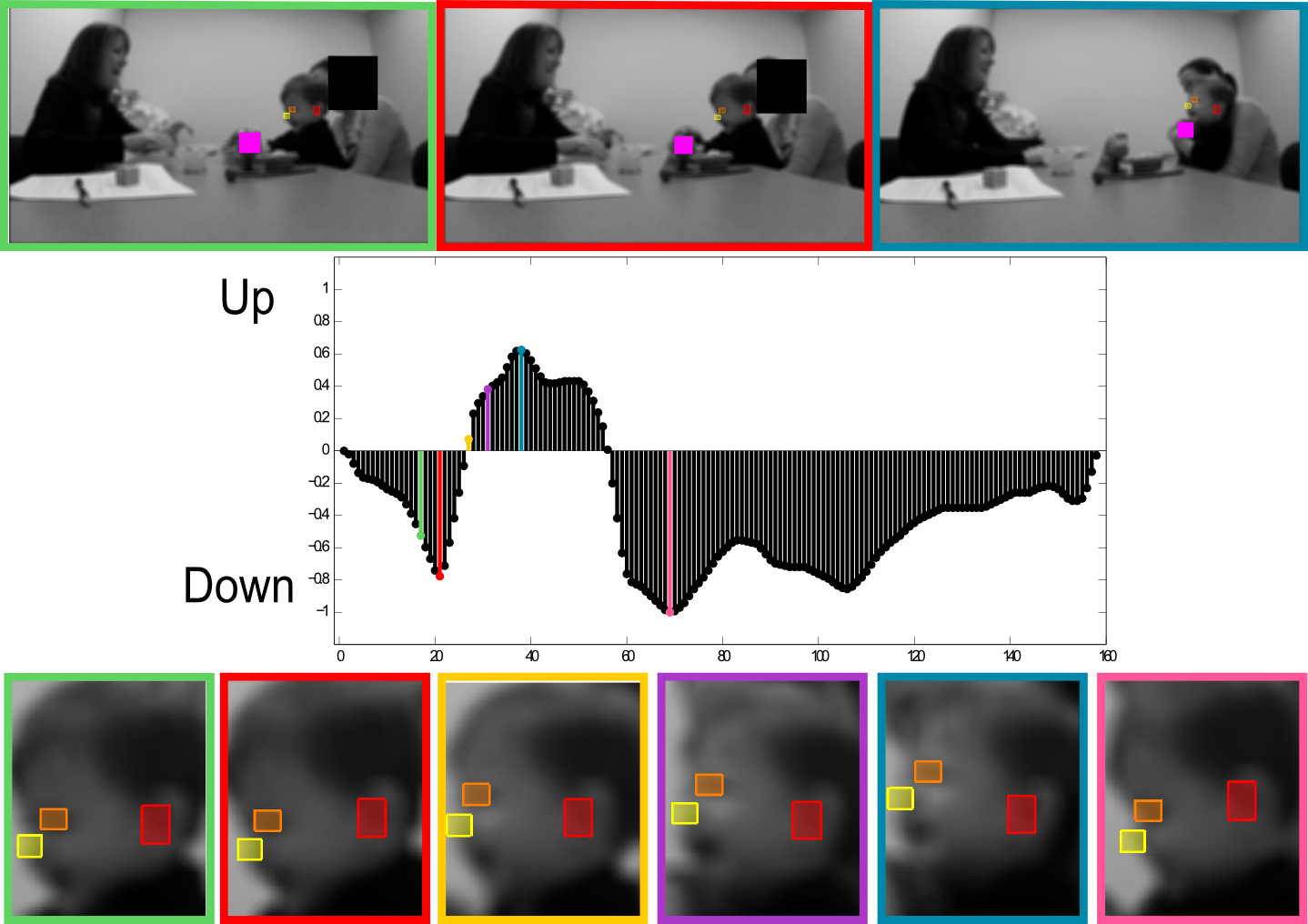}

\vskip.5cm

\includegraphics[width=.8\textwidth]{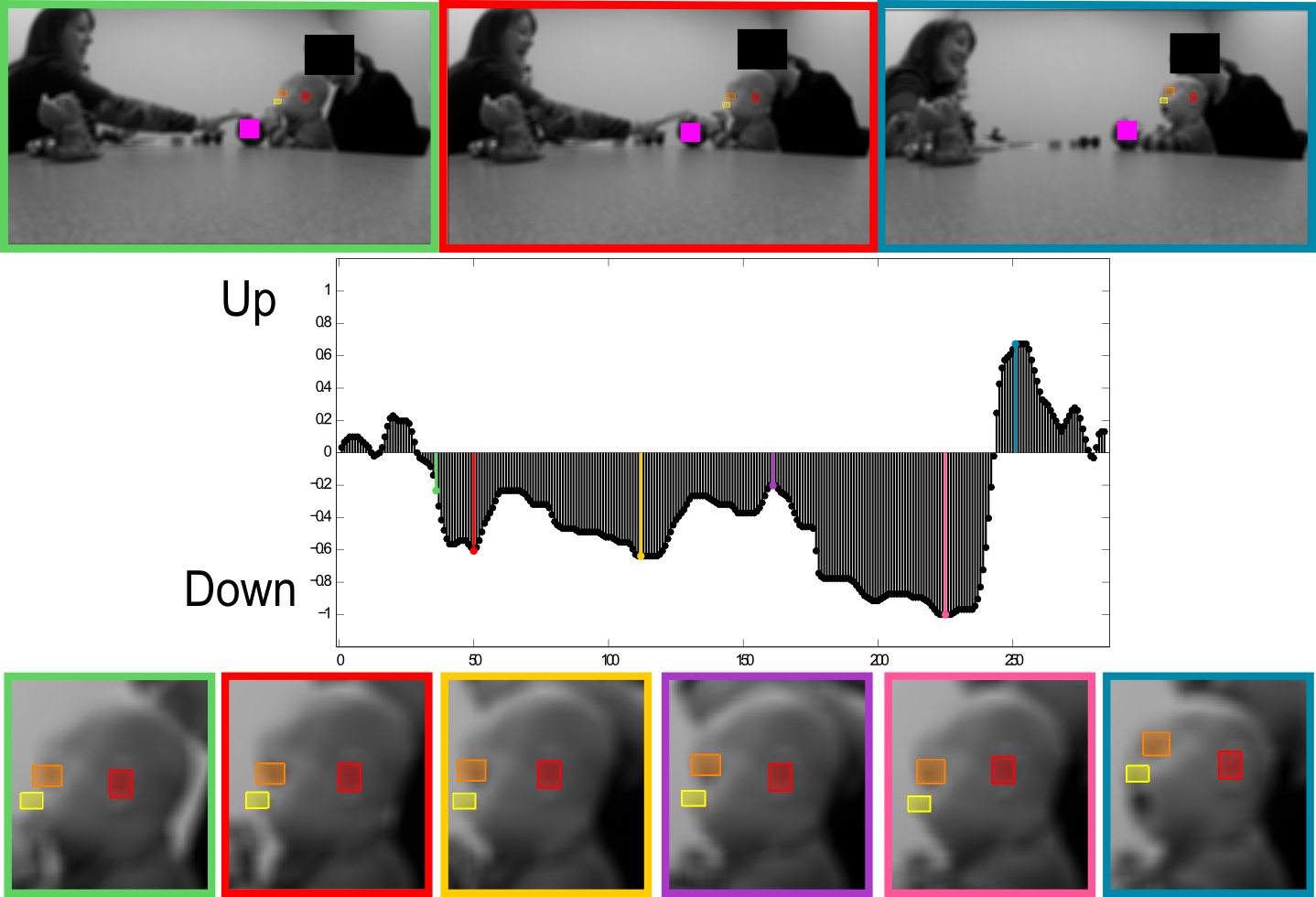}

\caption{Ball rolling activity. \textbf{Top:} when the ball contacts the child, when the child looks down at the ball, and when the child looks up at the clinician. \textbf{Middle: }
  changes in the pitch motion ($y$-axis) for each frame ($x$-axis). The dotted line represents when the ball contacts the participant.
  \textbf{Bottom:} 6 examples of the infant's face during the
  administration. All facial features are automatically detected and tracked.
  Colors identify corresponding images and spikes in the graph.
}
\label{f.shared_interest}
\end{figure*}

As stated before, the clinician's assessment of shared interest is partially done in the ball playing activity. Since the AOSI assessment of shared interest is done on a holistic scale, we provide examples of where our automatic method could prove useful. Figure~\ref{f.shared_interest} shows examples of our results from the ball playing activity during the freeplay session. For this particular activity, the clinician rolls a ball to the participant and analyzes if the participant shows shared interest. According to the AOSI, the participant shows shared interest if he/she either looks at the clinician or his/her caregiver after receiving the ball. Our automatic method is able to record and display the changes in the pitch motion of the participant. This allows the clinician to not only determine if the participant looked up after receiving the ball, but also how long it took him/her to look up and how long he/she became fixated on the ball. For both examples provided, the participants looked back up at the clinician after receiving the ball. In the first example, the participant looked up at the clinician within 0.73s (22 frames) of receiving the ball. On the other hand, it took the participant in the second example 7.17s (251 frames) to look up at the clinician after receiving the ball. Although each participant showed shared interest, the participant in the second example looks at the ball nearly 6.5s longer than the first participant before he/she looks back up at the clinician. By automating results such as these over large datasets, new patterns and risk behaviors could be established.

\subsection{Arm Asymmetry Analysis}
\label{ss.results-skeleton}

\begin{figure}
\centering
\includegraphics[width=\textwidth]{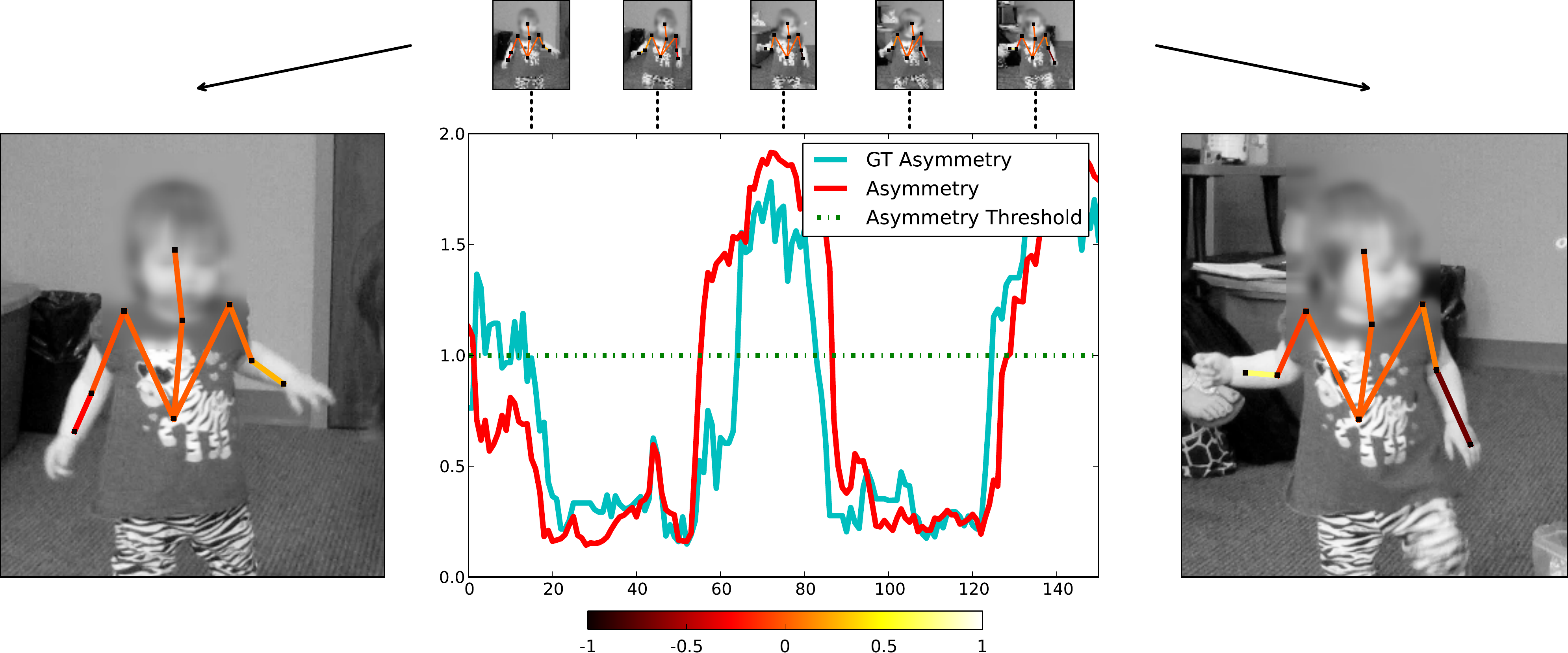}

\caption{Pose estimation performed for a video segment presenting
  participant \pnineteen{} walking unconstrained. We are interested in
  finding when the toddler is walking with asymmetric arm poses, a
  possible sign of ASD. We colorcode the upper arm with the
  corresponding asymmetry score $\ASu$ (see
  Appendix~\ref{ss.asymmetry-measures}) and the forearm using the
  final asymmetry score \ASa{}, after shifting the mean values to the
  interval $[-1,1]$ to denote the left/right arm segment with
  lowest/highest vertical coordinate. The graph depicts the absolute
  non-shifted final asymmetry score \ASa{} ($y$-axis) across time
  ($x$-axis), with $\ASa\geq1.0$ representing when the toddler's arms
  are not symmetric in the given frame. In this example, participant
  \pnineteen{} walks holding one forearm in (near) horizontal position
  pointing sideways, while extending the other arm downwards alongside
  her body (frames $0-18$, $63-85$, and $125-150$). We present the
  asymmetry scores obtained from the groundtruth skeleton in cyan in
  the graph. The asymmetry scores from the automatically computed
  skeleton and the ones obtained from the groundtruth skeleton
  correlate for this video segment, demonstrating the accuracy of the proposed technique. \label{f.p19-s01-asymmetry} }
\end{figure}

\begin{figure}
\begin{tabular}{cc}
\includegraphics[width=.45\textwidth]{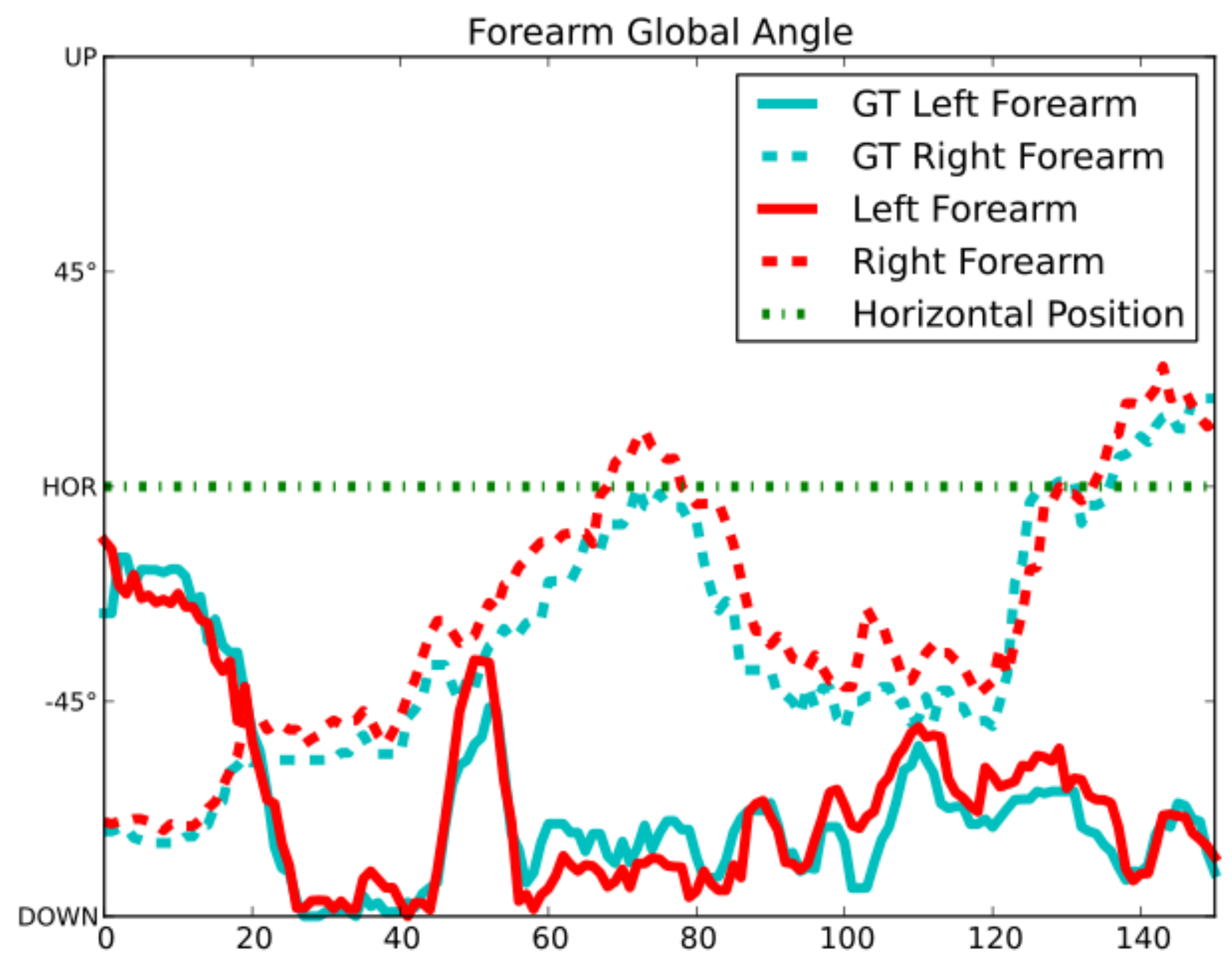} &
\includegraphics[width=.45\textwidth]{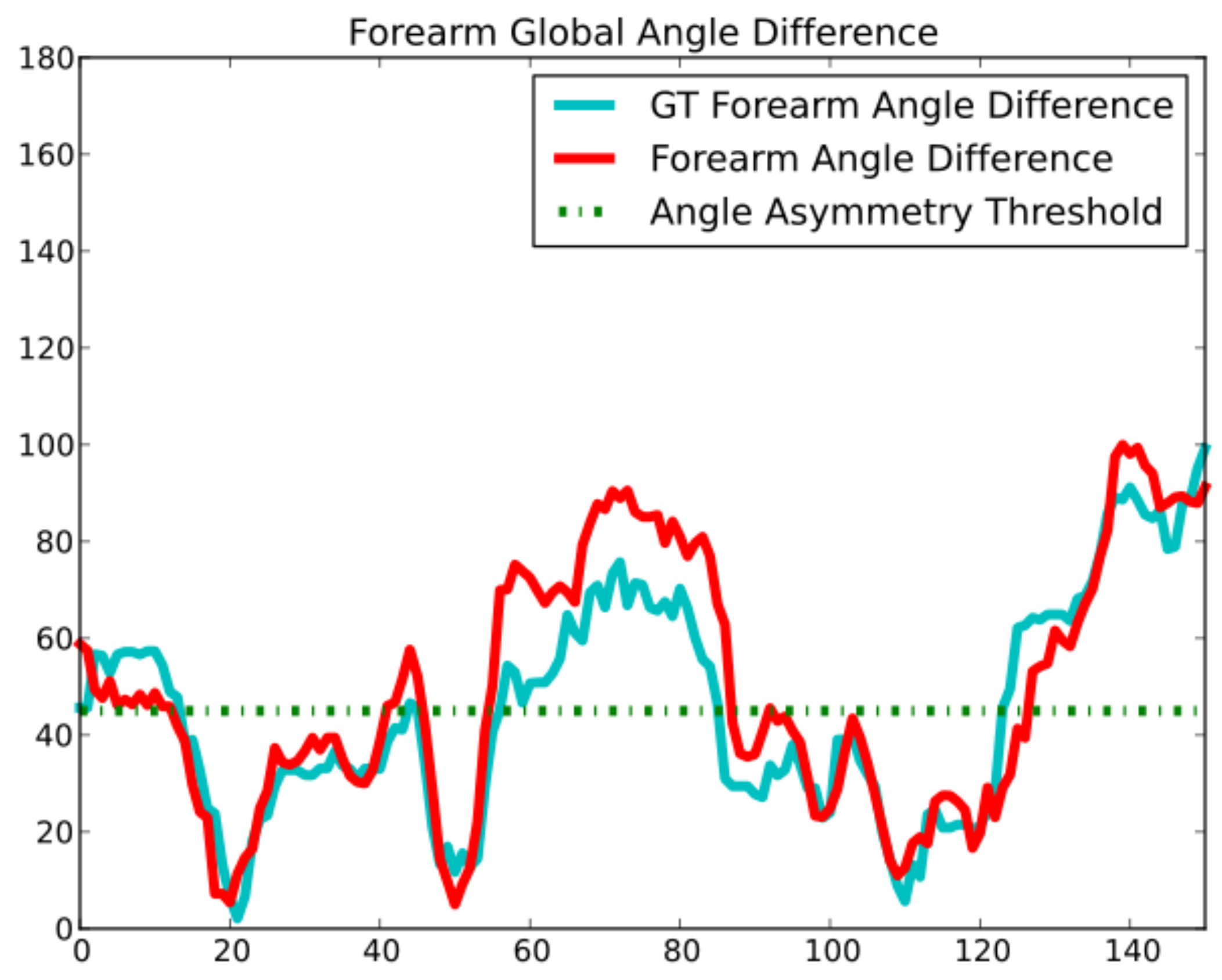}\\
(a) & (b)
\end{tabular}
\caption{The graphs in this figure depict other measures obtained from
  the skeletons from the video segment in
  Figure~\ref{f.p19-s01-asymmetry}. Namely, (a) the 2D global angle
  values for participant \pnineteen's left and right forearms; (b) the
  difference \ADf{} between the corresponding angles. Those measures
  essentially indicate where each forearm is pointing to w.r.t. the
  horizontal axis (up, down, horizontally). Among other things, these
  angles might also indicate asymmetry when
  $\ADf\geq45^o$~\citep{Hashemi12}. Since we compute a 2D skeleton,
  false positives/negatives might occur due to off-plae rotations.
  By analyzing both \ADf{} and \ASa{} from
  Figure~\ref{f.p19-s01-asymmetry}, one can often rule out false
  positives/negatives that occur (e.g., the false negative indication
  of asymmetry between frames $0$ and $18$ by the \ASa{} graph in
  Figure~\ref{f.p19-s01-asymmetry} is captured by the \ADf{} graph in
  (b)).
\label{f.p19-s01-angles}}
\end{figure}

\begin{figure}
\centering
\includegraphics[width=\textwidth]{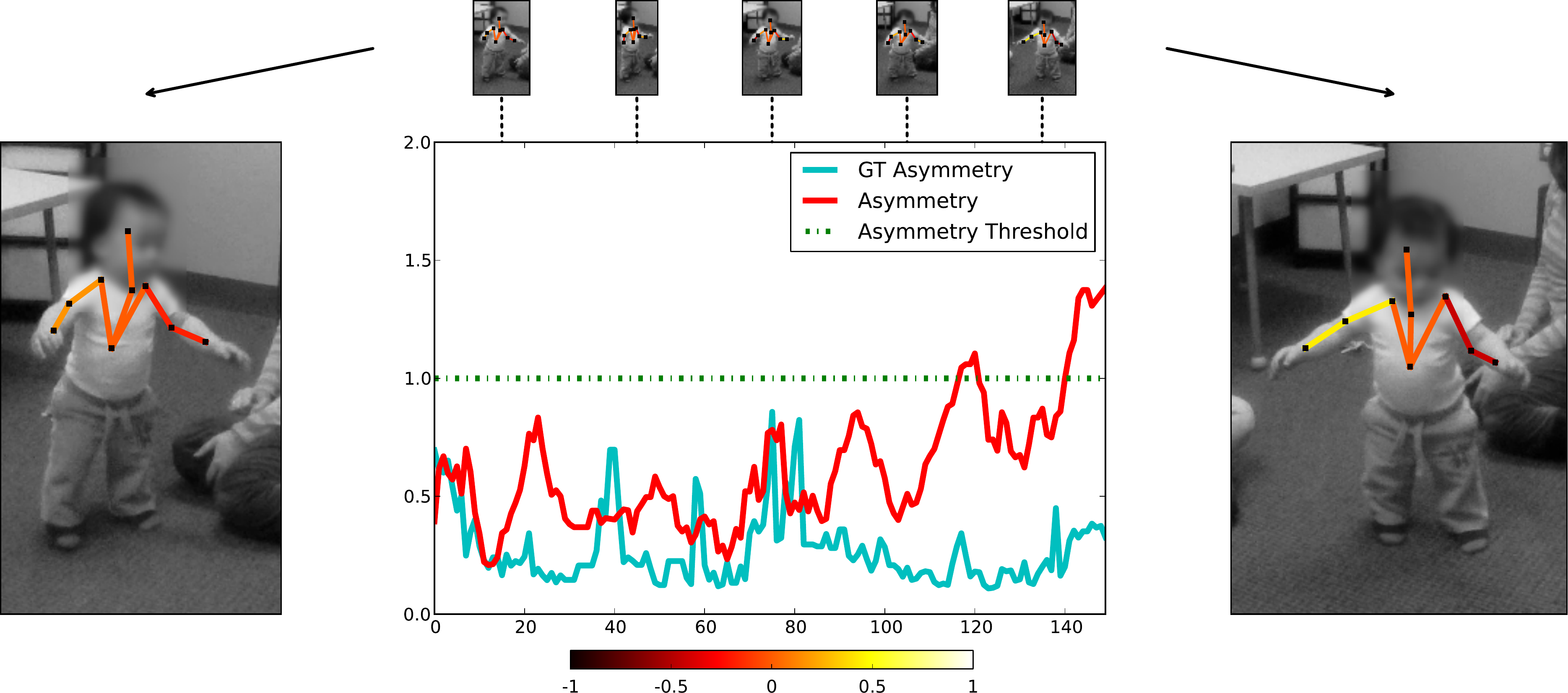}

\caption{This graph represents a video segment from participant
  \pfive. In this case, the opposite situation that occurs with the
  video segment of participant \pnineteen~from
  figures~\ref{f.p19-s01-asymmetry} and~\ref{f.p19-s01-angles} is
  happening. Namely, the asymmetry scores $AS^*$ between frames
  $20-80$ denote symmetric behavior for both the groundtruth and our
  automatically computed skeleton, while the \ADf{} values in
  Figure~\ref{f.p05-s01-angles} (b) indicate false positive
  asymmetry. Such disagreement is due to participant \pfive~walking in
  near frontal view with his arms wide open. Hence, the stickman's
  left forearm appears in horizontal position, while the stickman's
  right forearm points vertically down (i.e., $\ADf\geq60^o$ for the
  better part of frames $20-80$). Such situation shows the importance
  of considering multiple asymmetry measures to overcome the
  shortcomings of using the 2D skeleton under projective
  transformations.
\label{f.p05-s01-asymmetry}}
\end{figure}

\begin{figure}
\begin{tabular}{cc}
\includegraphics[width=.45\textwidth]{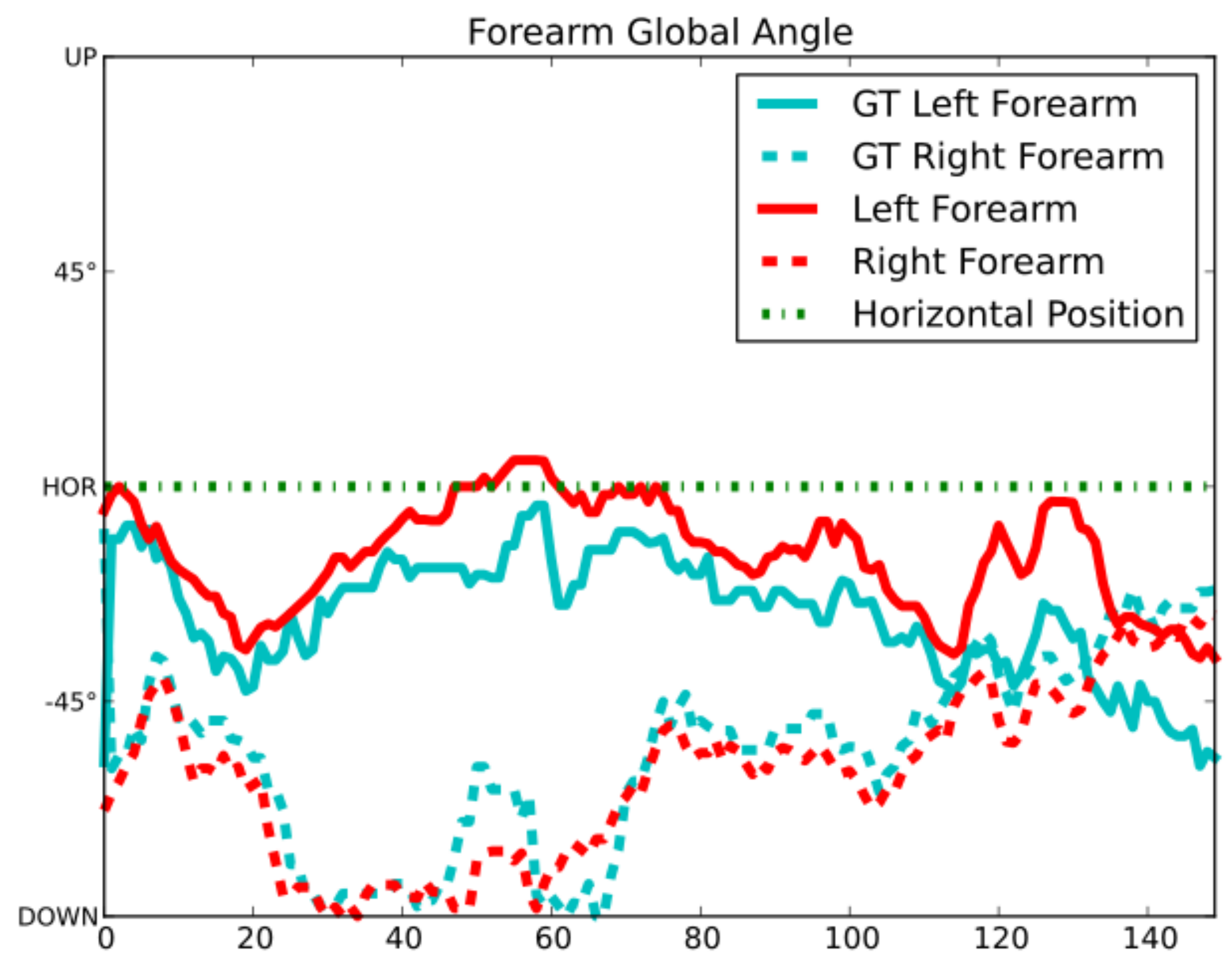}
&
\includegraphics[width=.45\textwidth]{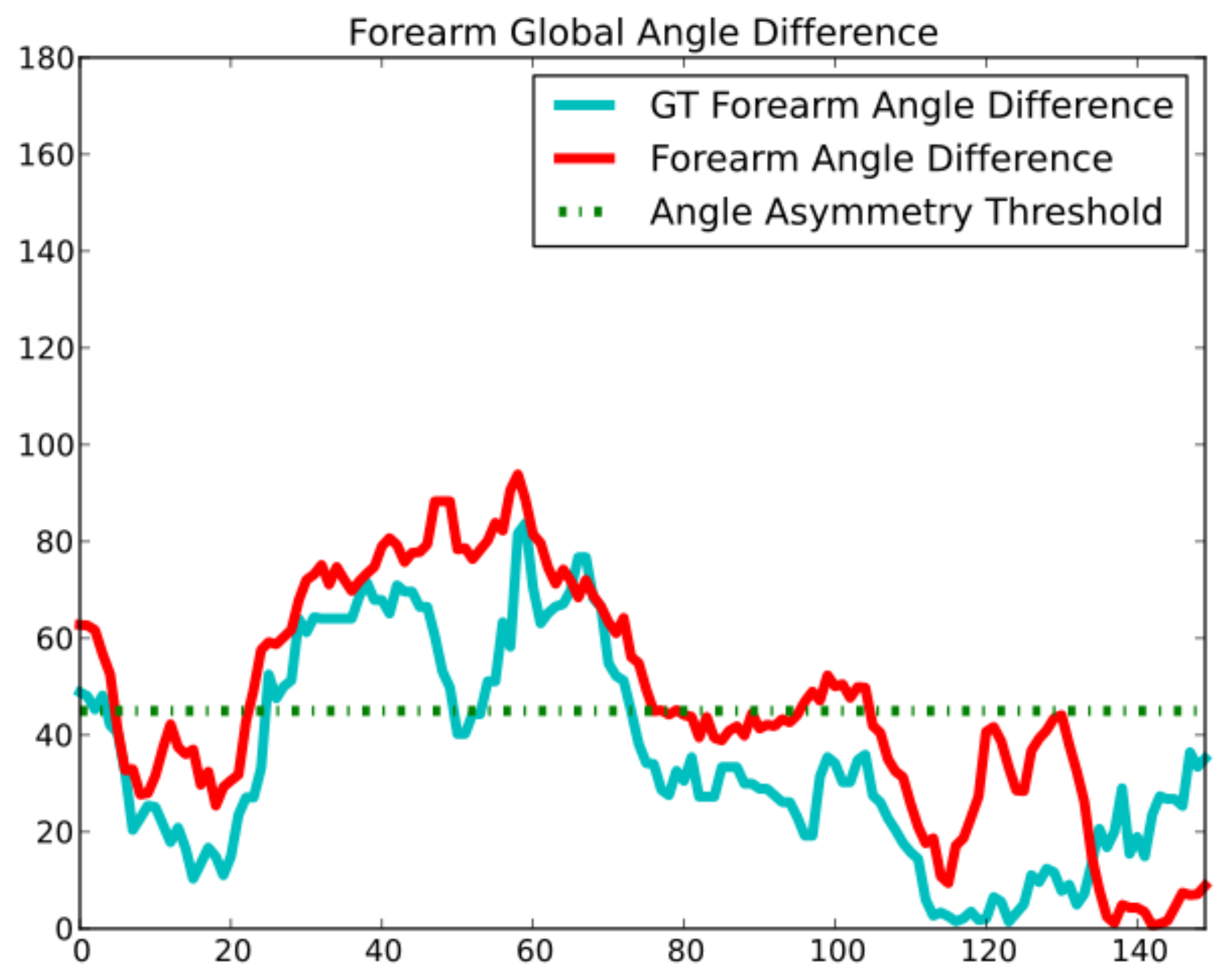}\\ 
(a) & (b)
\end{tabular}
\caption{Raw 2D global angles for the toddler's left and right
  forearms (a), and the corresponding angle difference \ADf{} (b) for
  the video segment of participant \pfive~in
  Figure~\ref{f.p05-s01-asymmetry}.\label{f.p05-s01-angles}}
\end{figure}

\begin{figure}
\centering
\includegraphics[width=\textwidth]{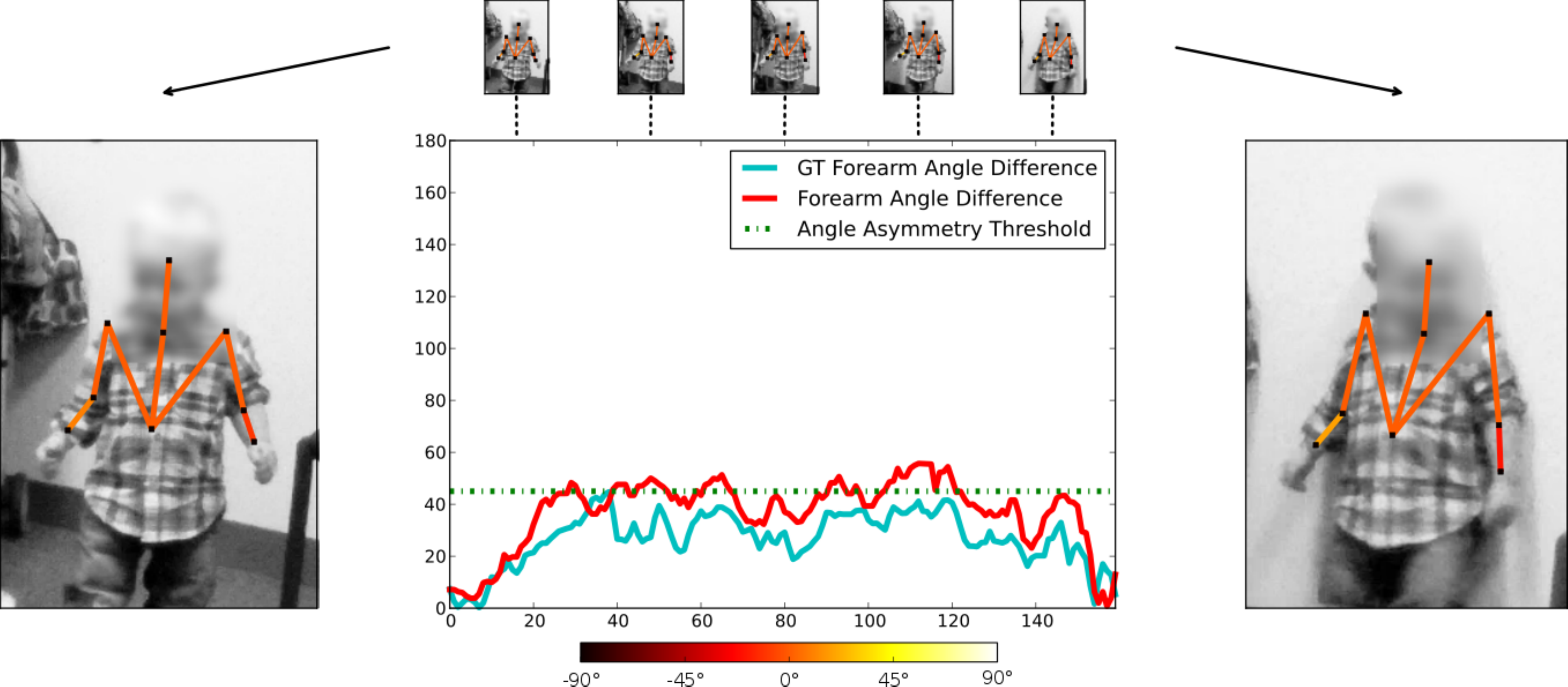}

\caption{First video segment of participant \pone, in which he walks
  with his arms mildly asymmetric. The graph depicts the asymmetry
  score \ADf, which reveals the behavior in some frames.
\label{f.p01-s01-angle-diff}}
\end{figure}

\begin{figure}
\centering
\includegraphics[width=\textwidth]{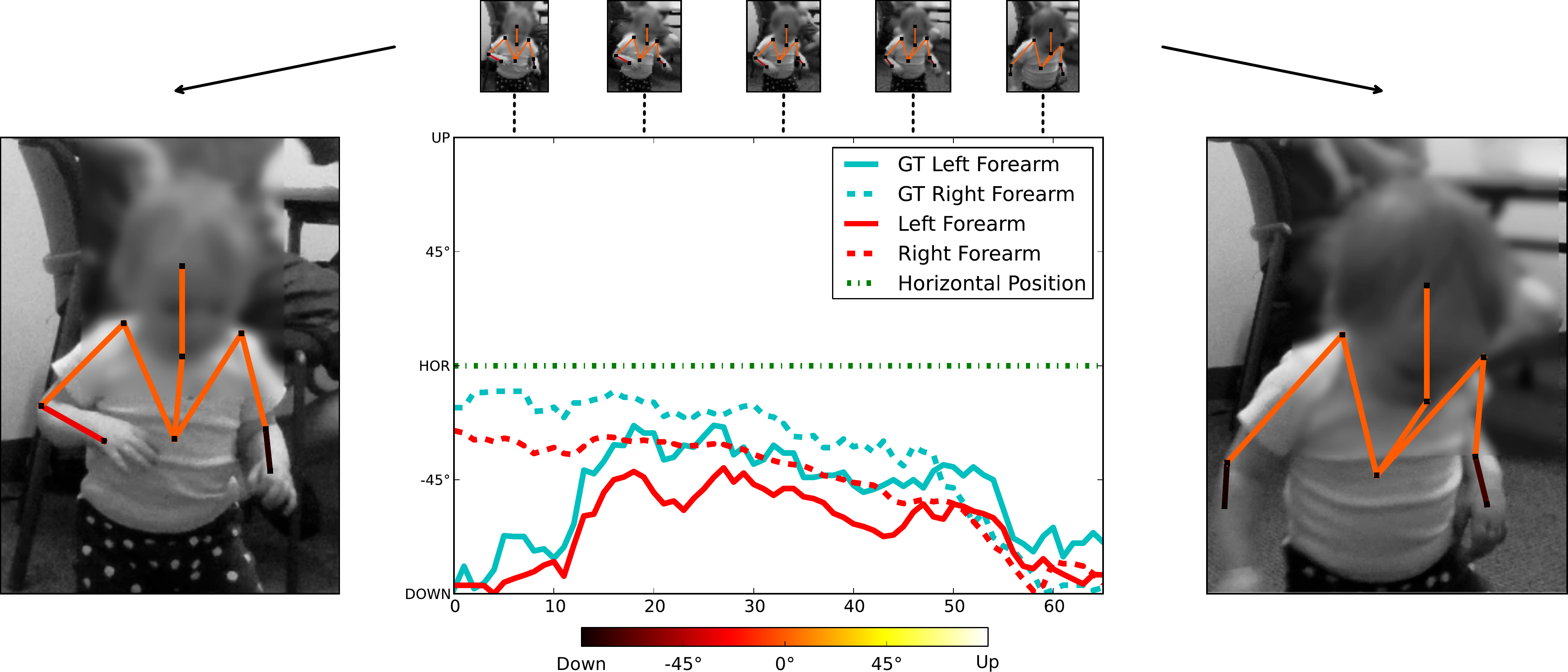}

\caption{First video segment of participant \pseven, in which she
  walks holding her arms parallel to the ground pointing forward. The
  graph depicts the forearm angles w.r.t. the horizontal axis. One can
  notice the aforementioned stereotypical motor pattern by analyzing
  from the graph that both forearms are close to the horizontal
  position for the better part of the video. This shows the array of
  stereotypical behaviors we may detect from our body pose estimation
  algorithm.
\label{f.p07-s01-angles}}
\end{figure}

\begin{figure}
\centering
\includegraphics[width=\textwidth]{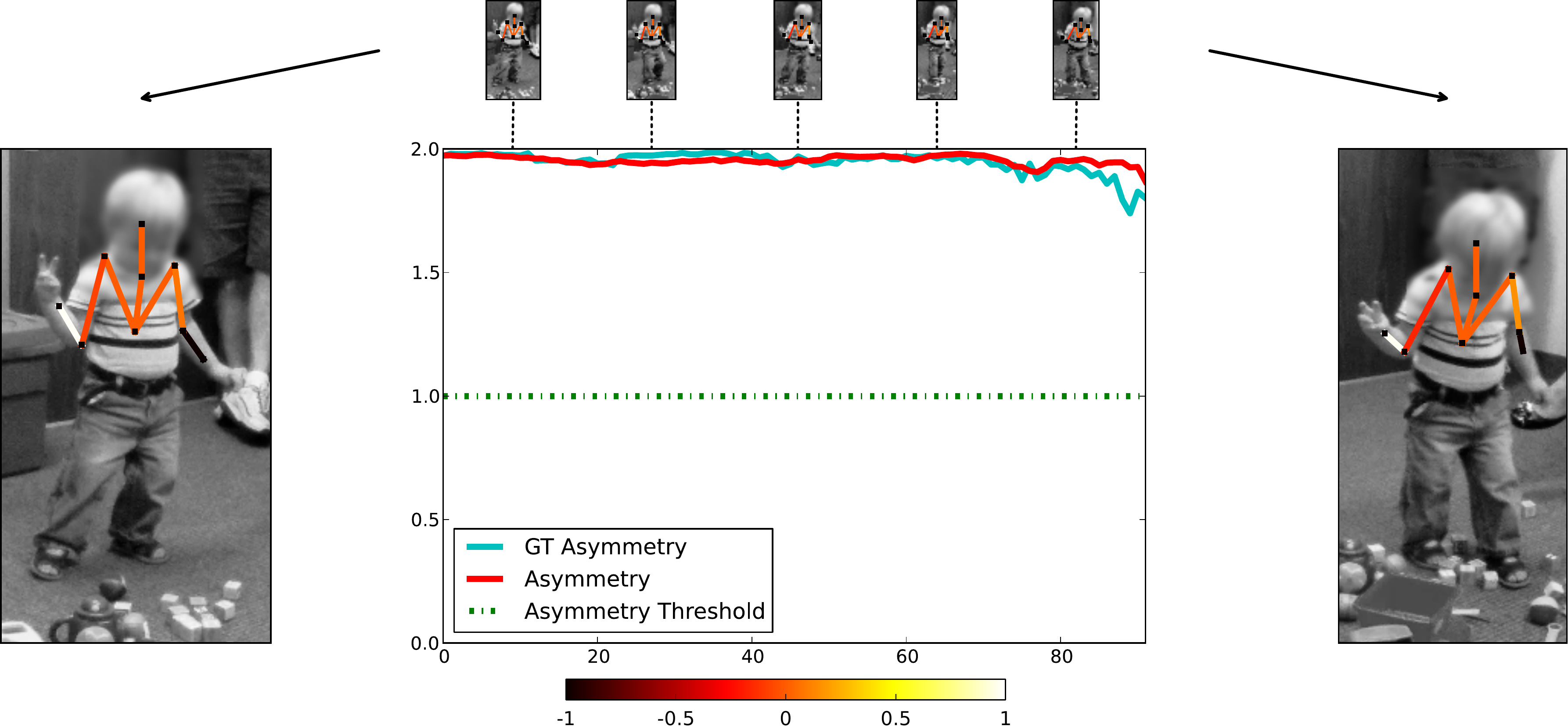}

\caption{First video segment of participant \ptwo{}. In this example,
  participant \ptwo{} is not only presenting asymmetric arm behavior
  throughout the entire video segment, but he is also presenting
  abnormal gait and hand behavior (other types of stereotypical motor
  behaviors). We intend to use the skeleton in the detection of such
  abnormal behaviors as well, by extracting different kinds of
  measures from it. Note that participant \ptwo{} has been diagnosed
  with autism.
\label{f.p02-s01-asymmetry}}
\end{figure}

\begin{figure}
\centering
\includegraphics[width=\textwidth]{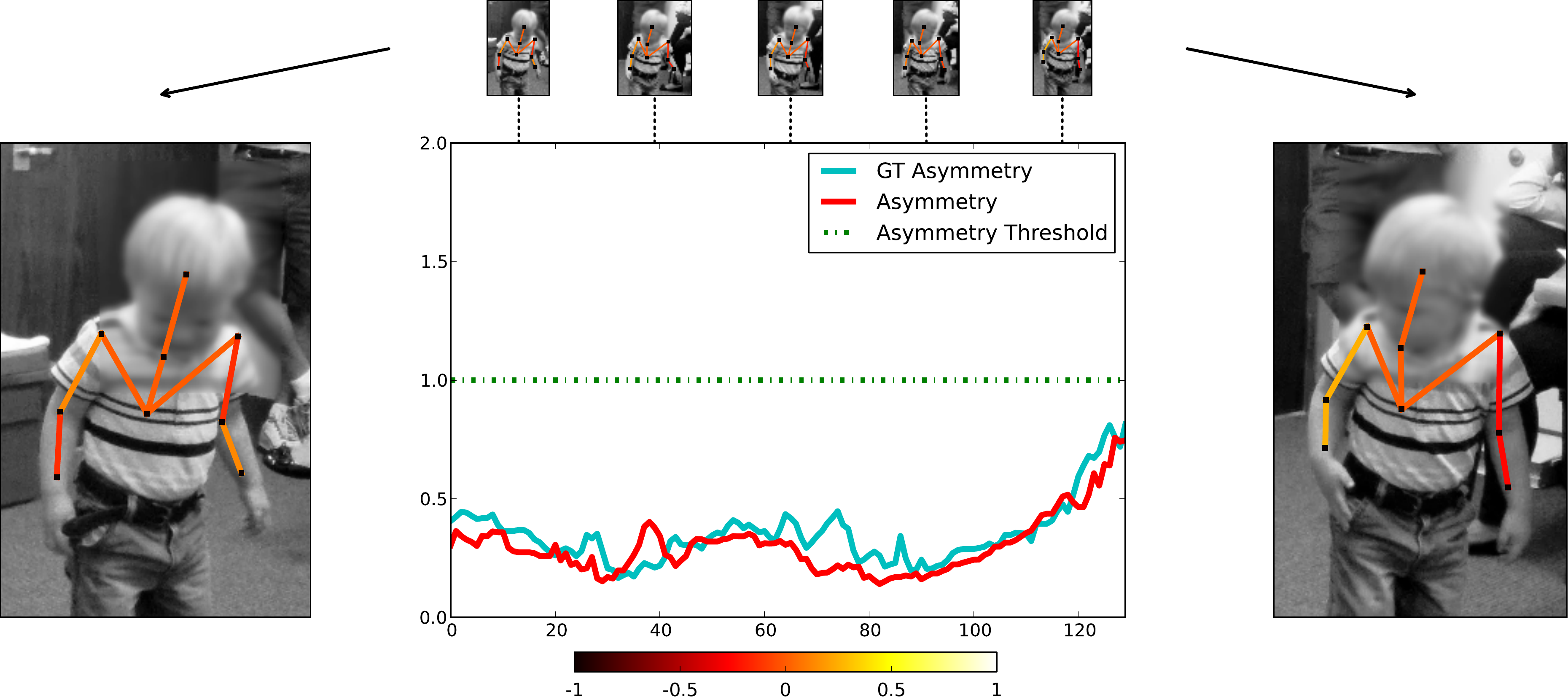}

\caption{The presented video segment and arm asymmetry graph indicate
  a moment in which participant \ptwo{} is walking symmetrically but
  toe-walking. Such indicative behaviors are paramount for early
  diagnosis, requiring constant monitoring and detection because their
  duration and intensity vary greatly among individuals.
\label{f.p02-s02-asymmetry}}
\end{figure}

We tested our body pose estimation algorithm in video clips in which
the entire body of the child can be seen~\citep[following][]{Esposito11}. We
compiled video sequences from ASD evaluation sessions of $6$ toddlers
(aged between $11$ to $16$ months), using one or two video segments to
ensure that each child was represented by one sequence with at least
$150$ frames in length ($5s$). For each video segment of every
sequence, a single segmentation mask was obtained interactively in the
initial frame~\citep{Spina11}. In contrast, \citet{Esposito11}
compiled $5$ minutes sequences at $8$ fps from $50$ participants, that
were manually annotated frame-by-frame using EWMN. Our participants
are fewer and our sequences shorter, though still sufficient, because
our dataset does not contain unsupported gait for longer periods; this
is in part because (1) not all participants from our dataset have
reached walking age and (2) the sessions took place in a small
cluttered room (left image in Figure~\ref{f.pose-estimation}). Hence,
we screened our dataset for video segments that better suited the
evaluation of our symmetry estimation algorithm, rather than
considering each child's case. We illustrate our results using six of
such video segments in
figures~\ref{f.p19-s01-asymmetry}-\ref{f.p02-s02-asymmetry} (with
groundtruth).

Since we are interested in providing measurements for the clinician,
the temporal graphs in
figures~\ref{f.p19-s01-asymmetry}-\ref{f.p02-s02-asymmetry} depict the
asymmetry score \ASa{}, the left and right forearms' global angles and
corresponding difference \ADf{}, for video segments of $5$
participants. Please refer to the captions of the aforementioned
figures for a discussion on the advantages and shortcomings of both \ASa{}
and \ADf{}. The forearms' global angles essentially denote where each
one is pointing to w.r.t. the horizontal axis (up, down,
horizontally). From these measurements, different data can be
extracted and interpreted by the specialists. \citet{Esposito11}, for
instance, look at two different types of symmetry: Static Symmetry
(SS) and Dynamic Symmetry (DS). The former assesses each frame
individually, while the latter evaluates groups of frames in a
half-second window. If at least one frame is asymmetric in a window,
then the entire half-second is considered asymmetric for DS. SS and DS
are then the percentage of asymmetric frames and windows in a video
sequence, respectively (the higher the number, the more asymmetrical
the pattern of walking).

Table~\ref{t.motor-pattern-eval} summarizes our findings for the $6$
participants. We adopt a strict policy by considering a frame
asymmetric only when both \ASa{} and \ADf{} agree (i.e., $\ASa\geq1.0$
and $\ADf\geq45^o$). Although we do not aim at fully reproducing the
work of~\citet{Esposito11}, we attempt to quantify asymmetry for each
video sequence by computing SS and DS according to our asymmetry
definition. The direct measures reflected by our temporal graphs for
each of the $6$ participants can be seen in our supplementary results,
along with all the videos.

\begin{table}
\centering
\caption{Symmetry data for the video sequences from $6$ different
  participants used in our experiments. We computed the Static
  Symmetry and Dynamic Symmetry~\citep{Esposito11} from the automatically obtained skeleton (Aut.), considering a frame
  asymmetric if both \ASa{} and \ADf{} agree (recall that the higher
  the number, the more asymmetrical the walking pattern). We also
  present the Static/Dynamic Symmetry values obtained from the
  groundtruth skeleton (GT), the clinician's evaluation about the video
  segments of each sequence, and the video sequence length. For the
  clinician's evaluation, we categorize the results as ``symmetric''
  (Sym), ``asymmetric'' (Asym), or ``abnormal'' (Abn --- i.e., some
  other stereotypical motor behavior is present on the video
  segment). \label{t.motor-pattern-eval} } {\footnotesize
\begin{tabularx}{\textwidth}{cYYYYYYc}
\toprule
\multirow{2}{*}{Part.} & \multicolumn{2}{c}{Static Sym.} & \multicolumn{2}{c}{Dynamic Sym.} & \multicolumn{2}{c}{Clinician's Seq. Eval.} & \multirow{2}{*}{Seq. Length (s.)} \\
\cmidrule(rl){2-3} \cmidrule(rl){4-5} \cmidrule(rl){6-7}
& Aut. (\%) & GT (\%) & Aut. (\%) & GT (\%) & Seg. $1$ & Seg. $2$ & \\
\midrule
\pnineteen & $36$ & $34$ & $64$ & $55$ & Asym & - & $5.0$\\
\pfive & $0$ & $0$ & $0$ & $0$ & Sym & - & $5.0$  \\
\ptwo & $41$ & $41$ & $44$ & $44$ & Asym & Sym/Abn & $7.4$  \\
\pthree & $5$ & $0$ & $21$ & $0$ & Sym & Abn & $6.7$ \\
\pone & $0$ & $0$ & $0$ & $0$ & Asym & Sym & $7.6$ \\
\pseven & $29$ & $28$ & $36$ & $36$ & Abn & Abn & $6.5$ \\
\bottomrule
\end{tabularx}
}
\end{table}

Among the chosen participants, only participant \ptwo{} has been
diagnosed with autism at age of $18$ months. One of the video segments
we use clearly shows asymmetric arm behavior
(Figure~\ref{f.p02-s01-asymmetry}), as further confirmed by SS, DS,
and the clinician's evaluation in
Table~\ref{t.motor-pattern-eval}. However, such behavior is not a
direct example of asymmetry during regular walking pattern. It is
rather caused by different types of stereotypical behaviors (e.g.,
abnormal motor mannerism and ``clumsy'' gait), as revealed by
participant \ptwo's ADOS-T complex mannerism score of `$3$'
(`$0$'-`$3$' scale of increasing concern,~\citet{Luyster09}) and AOSI
atypical motor behavior score of `$2$' (binary scale using `$2$' for
\emph{atypical},~\citet{bryson2007prospective}). On the other chosen
segment that comprises the video sequence, his arms are symmetric even
though he is toe-walking. Still, if such behaviors can be captured by
our method, then more complex mannerisms can be addressed in the
future, beyond asymmetry detection (e.g., participant \ptwo{} also
presents frequent arm-and-hand flapping).

Participant \pnineteen{} has also presented asymmetric arm behavior
during regular walking pattern according to both our measurements
(figures~\ref{f.p19-s01-asymmetry} and~\ref{f.p19-s01-angles}) and the
clinician's assessment of the video sequence, even though her MSEL (Mullen Scales of Early Learning)
gross motor score was `$37$' (below
average,~\citet{Mullen95}). Conversely, participants \pthree{} and
\pone{} have presented predominantly symmetric arm behavior, even
though \pthree{} received an ADOS-T complex motor mannerism score of
`$2$' and \pone{} an AOSI atypical motor behavior score of `$2$.' Such
differences between our measurements and the assessment provided by
the clinical tools are probably due to other stereotypical motor
behaviors being detected throughout the evaluation session. That is,
gait symmetry is not an explicit item of either AOSI nor ADOS-T. On
the other hand, MSEL presents a more straightforward evaluation of
gross motor patterns which might be more correlated with gait
symmetry. The clinician assessed each child using several of these
clinical diagnostic tools at different time points. The clinician notes that
in the first video segment of participant \pone{} there might be some
arm asymmetry, while in the second video segment of participant
\pthree{} he walks with his forearms and hands parallel to the ground,
which could be a sign of bad gait. Our method deemed the first segment
of participant \pone{} symmetric mostly because only \ADf{} was able
to capture the asymmetry slightly, since it was mild
(Figure~\ref{f.p01-s01-angle-diff}). Hence, we could improve the
sensitivity of our measurements, at the cost of obtaining more false
positives, by relaxing our criterion and flagging frames if either
\ASa{} or \ADf{} point out asymmetry.

In some of the video segments we use, participants \pfive{} and
\pseven{} also walk while holding their forearms parallel to the
ground pointing forward. In the second video segment of participant
\pseven{}, this can be observed in the graph in
Figure~\ref{f.p07-s01-angles}, which shows that the forearms are in
near horizontal position thoughout the video segment. Thus, we might
also be able to detect those situations in the future from the
skeleton we automatically compute. As opposed to participant \pthree{}, participants
\pfive{} and \pseven{} might be holding their arms parallel to the
ground because they had just learned how to walk a couple of weeks
prior to the evaluation session. Only participant \pseven{} has shown
signs of concern, as well as higher asymmetry scores from our
measurements. In participant \pseven's video sequence, her unusual arm
position seems less natural than that of participant \pfive{}
(Figure~\ref{f.p05-s01-asymmetry}). Regardless, both participants have
obtained MSEL gross motor scores within the average range (`$58$' and `$57$,'
respectively).

Although our method agrees with the clinician's visual ratings about symmetry in $8$ out
of $10$ video sequences, pointing out when there is asymmetry and/or
some other atypical motor pattern, it is far from completely agreeing
with the clinical evaluation in every aspect about motor behavior (again, the expert's assessment is based on significantly more data). We
seek instead correlation between our results and the groundtruth
skeleton to aid in research and diagnosis by complementing human
judgement, since the latter will never be replaced. By analyzing our
graphs and Table~\ref{t.motor-pattern-eval}, one can notice that the
correlation exists. Thus, all affirmations previously stated are also
valid for the groundtruth symmetry measures. We have further shown
that our body pose estimation algorithm can be used to detect other potential stereotypical motor behaviors in the future, such as when the toddler
is holding his/her forearms parallel to the ground pointing
forward. Note that the behaviors here analyzed have only considered
simple measures obtained from the skeleton, whereas we can in the
future apply pattern classifiers to achieve greater discriminative
power.

\section{Conclusion}
\label{sec:conclusion}

This work is the first achieved milestone in a long-term project for
non-invasive early observation of children in order to aid in diagnosis of
neurodevelopmental disorders. With the goal of aiding and augmenting
the visual analysis capabilities in evaluation and developmental
monitoring of ASD, we proposed (semi-)automatic computer vision tools to
observe specific behaviors related to ASD elicited during AOSI,
providing both new challenges and opportunities in video analysis. The
proposed tools significantly reduce the effort to only requiring
interactive initialization in a single frame. We focused on four
activities performed during the battery of assessments of development
and behaviors related to ASD: three activities were performed during
the AOSI and were related to visual attention and one which involves
motor patterns observed at any point during the assessment process. We
developed specific algorithms for these activities,
obtaining a clinically satisfactory result.

The improvement and extension of the proposed methods is an ongoing
work. For the visual attention tests, we plan on complementing the
estimation of the child's motions with estimating the clinician's
movements in order to correlate both. For the assessment of the motor
patterns, we will incorporate 3D information using a richer 3D human
model. Note that our body pose estimation method can be done in fully
automatic fashion by learning the CSM from a sufficiently large
training dataset and applying it for single image body pose estimation
(similarly to the work of~\citet{Zuffi12}). Also, there is no need for
a human intervention in video segment selection, these are easily
identified either by time coding, voice commands, or automatically
finding the objects (e.g., the ball). Of course, there are additional
behavioral red flags of ASD, both included in and beyond the scope of
AOSI, which we aim at addressing in the future.
An interesting future direction would be to use our symmetry measurements to identify real complex motor mannerisms from more typical toddler movements.\footnote{Bilateral and synchronized arm flapping is common in toddlers as they begin to babble, being hard to judge whether this is part of normal development or an unusual behavior. This issue clearly applies to \pfive's and \pseven's clips from their 12-month assessments.}
This extension also includes
detecting ASD risk in ordinary classroom and home environments, a challenging task for which the developments here presented are a first step.

\appendix

\section{Tracking and Validating Facial Features} 
\label{app:tracking}

This section provides an overview of the technical aspects of the algorithm for tracking facial features and computing head motions from them. The large variability of the data and the lack of control about the camera positioning call for using very simple and robust features and algorithms.

We assume that, in the first frame, we have bounding boxes of three facial features: the left ear, left eye, and nose. To track these three facial features, and following a scheme loosely based on the KLD tracker~\citep{Kalal}, we use dense motion estimation coupled with a validation step that employs an offline-trained facial feature detector. The dense motion estimator~\citep{Tepper12} tracks the features with high accuracy in most cases, but when the child's head moves quickly, illumination changes can sometimes cause the tracker to lag behind the features. Thus we validate the output of the tracker using facial feature detectors in every frame.

To validate the features we train left eye, right eye, left ear, and
 nose detectors based on the method proposed by~\citet{Dalal} \citep[see
 also][]{Everingham06}. Our method uses multiscale Histograms of
 Orientated Gradients (HOG) as descriptors to represent each facial
 feature, and then classifies these descriptors using a Support Vector
 Machine. As positive training samples, we use hand labeled facial
 patches from children in our experimental environment. As negative
 training samples, we extract random patches from around multiple
 children's faces.

 For each frame, search areas for the facial feature detectors are
 defined around the bounding boxes given by the tracker. Since the left
 eye, left ear, and nose are present in every frame for the given
 camera position, we impose a lenient classifier threshold and
 geometrical constraints (e.g., the left eye must be higher and to the
 left of the nose). The tracker's bounding boxes are validated if their
 centers are within the bounding boxes returned by the detectors;
 however, if the tracker's centers are outside of the detector's
 bounding boxes for two consecutive frames, then the corresponding
 bounding box for the tracker is reset to a new location within the
 detector's bounding box. Determining the presence of the right eye
 aids in the estimation of the yaw motion. The search area for the
 right eye, which is not tracked since it appears and disappears
 constantly, is based on the locations of the
 detected left eye and nose.

 \subsection{Yaw and Pitch Motion Estimation from Facial Features}

 In our setup, the child's face is predominantly in a profile view for the
 Sharing Interest activity. As a way to provide an accurate motion
 estimation of the pitch angle we cumulatively sum the vertical
 coordinate changes of the left eye and nose with respect to the left
 ear across a period of $2$ frames. We expect a positive sum when the
 child is looking up and a negative sum when the child is looking down,
 the magnitude representing how much the child is looking up or down.

\begin{figure*}[b]
 \centering
 \includegraphics[width=.7\textwidth]{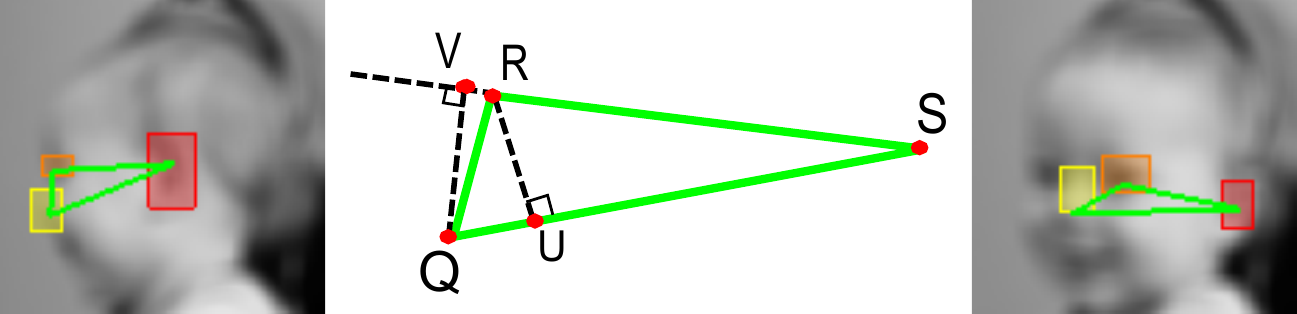}
 \caption{The triangle created by the left ear, left eye,
   and nose. The leftmost and rightmost images depict the triangle when
   the infant is looking right and more towards the camera, respectively. The middle image shows the points used
   for calculating $\widehat{\mathrm{yaw}}$.
 }
 \label{f.triangle}
 \end{figure*}

 For estimating the yaw angle motion in the Visual Tracking and
 Disengagement of Attention activities, we calculate two ratios based on the
 triangle created by the left ear, left eye, and nose
 (Figure~\ref{f.triangle}); we also use information about the presence
 of the right eye. Let $Q$, $R$, and $S$ denote the locations of the
 nose, left eye, and left ear, respectively. For the first ratio
 $r_{\text{NoseToEye}}$, we project $R$ into the line defined by $QS$,
 thus defining the point $U$; we then define $r_{\text{NoseToEye}} =
 |US|/|QS|$, where \mbox{$|\cdot|$} is the Euclidian distance. For the second
 ratio we project $Q$ into the line defined by $RS$, defining
 $r_{\text{EyeToEar}} = |VR| / |RS|$.

 The two ratios $r_{\text{EyeToEar}}$ and $r_{\text{NoseToEye}}$ are
 inversely proportional. Looking at Figure~\ref{f.triangle} we can
 observe that when the face is looking in profile view,
 $r_{\text{EyeToEar}}$ will be large and $r_{\text{NoseToEye}}$ will be
 small; conversely when the face is in frontal view (looking more towards
 the camera). To combine these two ratios into one value, we calculate
 the normalized difference between them,
 $\widehat{\mathrm{yaw}}=\frac{r_{\text{EyeToEar}}-r_{\text{NoseToEye}}}{r_{\text{EyeToEar}}+r_{\text{NoseToEye}}}$. Thus,
 as the child is looking to his/her left, $\widehat{\mathrm{yaw}}$ goes to -1; and
 as the child is looking to his/her right, $\widehat{\mathrm{yaw}}$ goes to 1. The
 presence of the right eye
 further verifies that the infant is looking
 left.
 We incorporate if the right eye is present or not to verify that the
 infant is looking left or right at the maximum and minimum $\widehat{\mathrm{yaw}}$
 values.

\section{Body pose estimation using the Object Cloud Model}
\label{app:bodyPose}

The Object Cloud Model (OCM) is represented by a \emph{fuzzy object}
(\emph{cloud image}) in which each pixel receives one out of three
possible values: object, background, or
uncertainty~\citep{Miranda10-TR-IC}. The silhouette variations are
captured by the uncertainty region, which represents the area where
the real object's boundary is expected to be in a new test image
(Figure~\ref{f.pose-estimation}). OCM then treats the object detection
task (locating the object of interest in an image) and delineation
(defining the object's spatial extent) in a synergistic
fashion. Namely, for each possible object position in an image
(frame), OCM executes a delineation algorithm in the uncertainty
region and evaluates if the resulting segmentation mask yields a
maximum score for a given search criterion. This maximum should be
reached when the uncertainty region is properly positioned over the
real object's boundary. Ideally, if the uncertainty region is well
adapted to the object's new silhouette and the delineation is
successful, the object search is reduced to translating the model over
the image.

When the object is composed of multiple correlated substructures, such
as the parts of the human brain, a Cloud System Model (CSM) may be
created by transforming each substructure into an OCM and taking into
account the relative position between them during the
search~\citep{Miranda10-TR-IC}. We consider the human body as the
object of interest, divide it into each of its major structures
(torso, head, arms, and legs), and connect those structures using a 2D
stickman model to create a CSM in a given initial frame
(figures~\ref{f.pose-estimation}\subref{f.pose-estimation1_seg}-\subref{f.pose-estimation1_stick}). Then,
the resulting CSM is used to automatically find the toddler's body
frame-by-frame in the video segment
(figures~\ref{f.pose-estimation}\subref{f.pose-estimation2_seg}-\subref{f.pose-estimation2_stick}).

We require a single segmentation mask, obtained interactively in the
first frame~\citep{Spina11}, to compute the model. Then, the body pose
search maximizes the search criterion by applying affine transforms to
each CSM cloud, respecting the body's tree hierarchy (rooted at the
torso), until the model finds the new pose. We use dense optical flow~\citep{Tepper12}
to reinitialize the pose search for the next frame, which is repeated
until the end of the video segment. 
details about creating and using CSM for 2D body pose estimation).
If necessary, one may correct the pose search by stopping our method and
providing a new segmentation mask in a given frame. This is a standard procedure in popular video analysis packages such as Adobe's {\it After Effects.}

\newcommand{\figheight}{0.15\textheight}

\begin{figure*}[b]

\centerline{
\hfill
\includegraphics[height=\figheight]{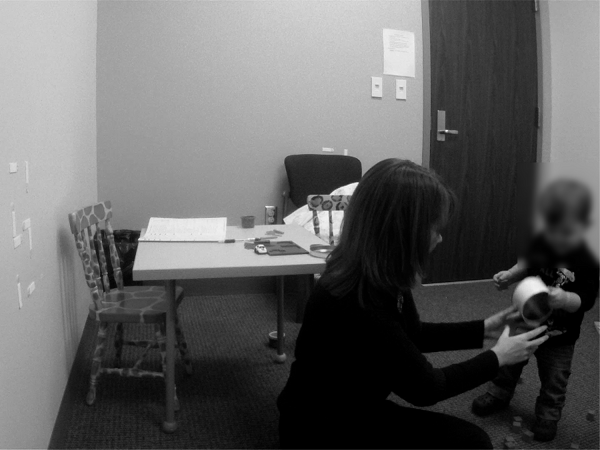}
\hfill
\subfloat[]{
    \includegraphics[height=\figheight]{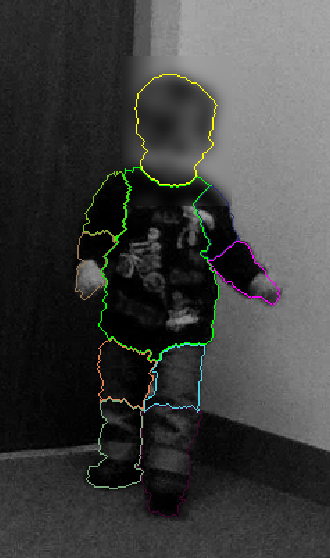}
    \label{f.pose-estimation1_seg}
}
\hfill
\subfloat[]{
  \includegraphics[height=\figheight]{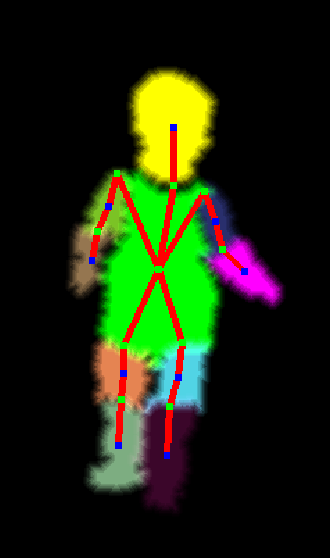}
    \label{f.pose-estimation1_stick}
}
\hfill
\subfloat[]{
    \includegraphics[height=\figheight]{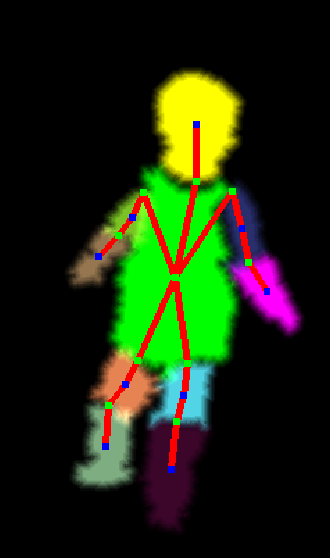}
    \label{f.pose-estimation2_seg}
}
\hfill
\subfloat[]{
  \includegraphics[height=\figheight]{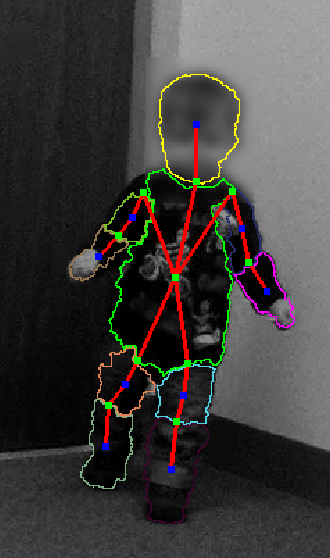}
    \label{f.pose-estimation2_stick}
}
\hfill
}

\caption{{\bf Left:} General scene capturing the ASD evaluation
  session. {\bf Right:} Overall segmentation and position tracking
  scheme. \protect\subref{f.pose-estimation1_seg} Segmentation mask
  $L^t$ provided at an initial frame
  $t=0$. \protect\subref{f.pose-estimation1_stick} CSM computed from
  $L^t$ and the 2D stickman used to connect the clouds corresponding
  to each body part. \protect\subref{f.pose-estimation2_seg}
  Transformed CSM at frame
  $10$. \protect\subref{f.pose-estimation2_stick} Segmentation and
  final pose estimation.  }
\label{f.pose-estimation}
\end{figure*}

\subsection{Arm Asymmetry Score Computation}
\label{ss.asymmetry-measures}

For computing the asymmetry score \ASa{} from the estimated skeleton,
we first define the following normalized asymmetry score for each arm
segment:
\begin{eqnarray}
AS(\alpha) & = & \frac{2.0}{1.0+\exp{(-\frac{\alpha - \tau}{\sigma_{\tau}}})}
\label{e.seg-asymmetry-score},
\end{eqnarray}
where $\alpha$ is the absolute difference between either global or
relative 2D angles obtained from corresponding left/right arm
segments, $\tau$ is a given asymmetry threshold, and $\sigma_\tau$ is
a parameter set to control acceptable asymmetry values. Considering
EWMN's accuracy, we set the asymmetry threshold $\tau=45^o$. We have
empirically observed that $\sigma_\tau = \frac{\tau}{3}$ helps coping
with near asymmetrical poses when outputing the asymmetry score.

\begin{figure}
\begin{center}
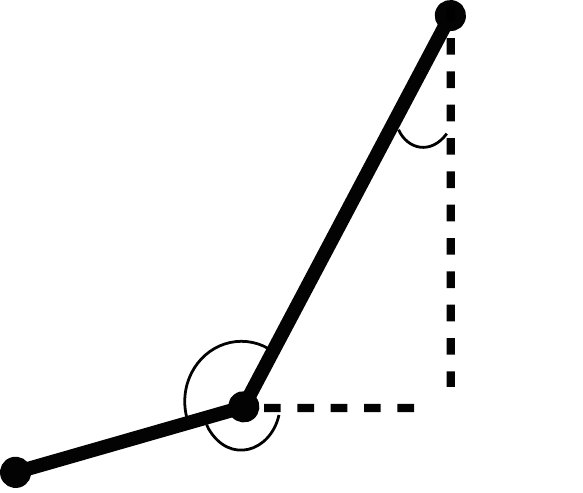
\end{center}
\caption{Angles used to compute arm asymmetry. The upper arm asymmetry
  score $\ASu$ considers the global angle $\hat{u}$ with respect to
  the vertical axis. The forearm asymmetry score $\ASf$ considers the
  relative angle $\hat{e}$ formed by the forearm and the upper arm at
  the elbow. The absolute angle difference between corresponding
  left/right arm segments is used in
  Equation~\ref{e.seg-asymmetry-score} as $\alpha$ to output an
  asymmetry score for each arm segment, the maximum representing the
  overall arm asymmetry score $\ASa$. Global angle $\hat{f}$ defines
  where the forearm is point to w.r.t. the horizontal axis
  (up/down/horizontally), while the corresponding absolute difference
  $\ADf$ between left and right forearm angles is used as a another
  asymmetry measure.\label{f.as-angles}}
\end{figure}

For the upper arm asymmetry score $\ASu$, we set $\ASu=AS(\alpha_u)$
in Equation~\ref{e.seg-asymmetry-score} with
$\alpha_u=|\hat{u}_l-\hat{u}_r|$ being the absolute difference between
the global angles $\hat{u}_l$ and $\hat{u}_r$ formed by the left and
right upper arms with the vertical axis, respectively
(Figure~\ref{f.as-angles}). The forearm asymmetry score
$\ASf=AS(\alpha_f)$ is similarly defined by setting
$\alpha_f=|\hat{e}_l - \hat{e}_r|$, where $\hat{e}$ is the relative
forearm angle with respect to the upper arm formed by the elbow
(Figure~\ref{f.as-angles}). To ensure that we are considering
symmetric arm poses, we mirror the skeleton on the vertical axis to
compute all global angles on the $1^o$ and $4^o$ quadrants of the
cartesian plane. The asymmetry score \ASa{} for the entire arm is
finally defined as
\begin{eqnarray}
  \ASa & = & \max{\{\ASu,\ASf\}}.  \label{e.asymmetry-score}
\end{eqnarray}

The rationale behind Equation~\ref{e.asymmetry-score} is that if the
toddler's upper arms are pointing to different (mirrored) directions,
then the arms are probably asymmetric and $\ASu$ should be high (i.e.,
$\ASa\geq1.0$). Otherwise, if $\ASf$ is great then one arm is probably
stretched while the other one is not, thus suggesting arm
asymmetry. Regardless, we may also use where the forearms are
pointing to as another asymmetry measure, by analysing their global
mirrored angles $\hat{f}_l$ and $\hat{f}_r$ w.r.t. the horizontal axis
(Figure~\ref{f.as-angles}). If the absolute difference
$\ADf=|\hat{f}_l-\hat{f}_r|$ between those global angles is greater
than $45^o$, for example, then the arm poses are probably
asymmetric~\citep{Hashemi12}.

\end{document}